\documentclass[10pt,twocolumn]{article}

\usepackage[a4paper,top=2cm,bottom=2cm,left=1.5cm,right=1.5cm,columnsep=0.35in]{geometry}

\usepackage[T1]{fontenc}
\usepackage[utf8]{inputenc}
\usepackage{libertine}
\usepackage{microtype}

\usepackage{graphicx}
\usepackage{xcolor}
\usepackage{amsmath}
\usepackage[libertine]{newtxmath}

\usepackage{booktabs}
\usepackage{tabularx}
\usepackage{array}
\usepackage{longtable}
\usepackage{multirow}
\usepackage{adjustbox}
\usepackage{rotating}
\usepackage{pdflscape}

\usepackage{float}
\usepackage{wrapfig}
\usepackage{caption}
\usepackage{subcaption}
\usepackage{afterpage}
\usepackage{placeins}
\usepackage{flushend}
\usepackage{cuted}   % provides strip environment for full-width content

\usepackage{enumitem}
\usepackage{titlesec}
\usepackage{abstract}
\usepackage{fancyhdr}
\usepackage{lastpage}
\usepackage{lettrine}
\usepackage{setspace}

\definecolor{darkblue}{RGB}{26,13,171}
\definecolor{accentblue}{RGB}{70,130,180}
\definecolor{lightgray}{RGB}{245,245,245}
\definecolor{mediumgray}{RGB}{120,120,120}
\definecolor{focusboxbg}{RGB}{252,252,254}
\definecolor{focusboxborder}{RGB}{180,190,210}
\definecolor{sectioncolor}{RGB}{20,50,100}
\definecolor{subsectioncolor}{RGB}{60,60,60}
\definecolor{stepcol}{RGB}{20,50,100}
\definecolor{substep}{RGB}{70,130,180}

\usepackage[square,numbers,sort&compress]{natbib}

\usepackage[breaklinks=true,colorlinks=true,linkcolor=darkblue,citecolor=darkblue,urlcolor=darkblue]{hyperref}
\usepackage{doi}         
\usepackage{cleveref}

\usepackage[most]{tcolorbox}
\tcbuselibrary{skins,breakable}

\usepackage{pifont}
\newcommand{\cmark}{\textcolor{green!60!black}{\ding{51}}}

\newcommand{\dash}{\textemdash}

\newcommand{\defnrow}[1]{%
  \multicolumn{13}{p{\dimexpr\textwidth-2\tabcolsep}}{\footnotesize\itshape #1}\\}
\newcommand{\gaprow}{\addlinespace[6pt]}

\titleformat{\section}
  {\normalfont\Large\bfseries\color{sectioncolor}}
  {\thesection}
  {0.8em}
  {}
  [\vspace{-0.3em}\color{accentblue}\rule{\columnwidth}{0.5pt}\vspace{-0.8em}]

\titleformat{\subsection}
  {\normalfont\large\bfseries\color{subsectioncolor}}
  {\thesubsection}
  {0.8em}
  {}
  [\vspace{-0.3em}]

\titleformat{\subsubsection}
  {\normalfont\normalsize\bfseries\itshape\color{subsectioncolor}}
  {\thesubsubsection}
  {0.8em}
  {}
  [\vspace{-0.3em}]

\renewenvironment{abstract}{%
  \begin{tcolorbox}[
    colback=lightgray,
    colframe=lightgray,
    boxrule=0pt,
    arc=2mm,
    left=8pt,
    right=8pt,
    top=8pt,
    bottom=8pt,
    before skip=0.5em,
    after skip=1em
  ]
  \noindent\small\textbf{\abstractname.}\space\ignorespaces
}{\end{tcolorbox}}

\newtcolorbox{focusbox}[2][]{%
  enhanced,
  breakable,
  colback=focusboxbg,
  colframe=focusboxborder,
  fonttitle=\bfseries\normalsize,
  coltitle=darkblue,
  colbacktitle=white,
  title={#2},
  boxrule=1pt,
  left=10pt, right=10pt,
  top=10pt, bottom=10pt,
  before skip=12pt,
  after skip=12pt,
  attach boxed title to top left={
     yshift=-2mm,
     xshift=10pt
  },
  boxed title style={
     colback=white,
     colframe=white,
     boxrule=0pt,
     sharp corners
  },
  #1
}

\newcolumntype{C}[1]{>{\centering\arraybackslash}p{#1}}
\newcolumntype{L}[1]{>{\raggedright\arraybackslash}p{#1}}

\begin{document}

% Title and authors
\twocolumn[
\begin{@twocolumnfalse}

\vspace*{1cm}

{\huge\bfseries From eye to AI: studying rodent social behavior in the era of machine learning\par}

\vspace{1cm}

{\large Giuseppe Chindemi$^1$, Camilla Bellone$^1$ \& Benoit Girard$^1$\par}

\vspace{0.3cm}

{\normalsize $^1$Department Basic Neuroscience, University of Geneva\par}

\vspace{0.3cm}

{\small Correspondence: camilla.bellone@unige.ch; benoit.girard@unige.ch\par}

\vspace{1cm}

\begin{abstract}
\noindent The study of rodent social behavior has shifted in the last years from relying on direct human observation to more nuanced approaches integrating computational methods in artificial intelligence (AI) and machine learning. While conventional approaches introduce bias and can fail to capture the complexity of rodent social interactions, modern approaches bridging computer vision, ethology and neuroscience provide more multifaceted insights into behavior which are particularly relevant to social neuroscience. Despite these benefits, the integration of AI into social behavior research also poses several challenges. Here we discuss the main steps involved and the tools available for analyzing rodent social behavior, examining their advantages and limitations. Additionally, we suggest practical solutions to address common hurdles, aiming to guide young researchers in adopting these methods and to stimulate further discussion among experts regarding the evolving requirements of these tools in scientific applications.
\end{abstract}


\vspace{1.1cm}

\end{@twocolumnfalse}
]

% Main content starts here
\section{Introduction}

\lettrine[lines=3]{S}ocial behavior is defined as the set of interactions between individuals of the same species, including forming groups, cooperating or competing for resources, and defending territory \cite{ref1}. These interactions are dynamic and reciprocal, and influenced by internal states like motivation, emotions, and past experiences, which are not directly observable. Critically, these internal states and behaviors are controlled by specific neural circuits and brain regions. For example, the hypothalamus, the ventral striatum, and prefrontal cortex have been implicated in mediating social behaviors, including aggression, cooperation, and social bonding \cite{ref2,ref3,ref4,ref5}. Although animal models have provided substantial insights into the neural basis of these behaviors, elucidating precisely how neural circuits control social interactions remains challenging. This requires advanced techniques, including simultaneous neural recordings, precise tracking of multiple animals, and detailed behavioral analyses, which often constrain experimental design due to their complexity.

Traditionally, neuroscience has primarily focused on simple social behaviors and tightly controlled experimental conditions \cite{ref6,ref7,ref8}. Analysis has primarily relied on basic statistical methods gathered through short-duration experiments and indirect observations, such as lever pressing, rearing, or time spent near social stimuli. While this reductionist approach has generated valuable insights into neurobiological functions \cite{ref9,ref10,ref11,ref12,ref13,ref14,ref15}, it does not capture the full complexity of social interactions. On the other hand, studies using free interaction paradigms typically quantify behaviors by frequency or duration \cite{ref2,ref16,ref17,ref18,ref19,ref20}, leaving subtle or unexpected behaviors unexamined due to methodological limitations including labor intensity, lack of scalability, and limited reproducibility.

As neuroscience embraces machine learning (ML), there is a significant opportunity to overcome these limitations with robust, scalable, and objective analytical methods. Studying social interactions with ML presents unique challenges distinct from single-animal analyses, such as tracking multiple animals simultaneously, defining interactive behaviors accurately, and capturing dynamic interactions. Consequently, ML methods must be specifically tailored for social contexts.

Historically, analysis of rodent social interaction relied heavily on human observation using ethograms, which provided standardized behavioral definitions \cite{ref21,ref22} (www.mousebehavior.org). Despite their value, these traditional methods faced significant limitations including subjectivity, low temporal precision, limited granularity, anthropomorphisms, and high labor demands \cite{ref23,ref24,ref25,ref26,ref27} (\hyperref[tab:human_annotation_limitations]{see Table 1}).

To address these issues, recent advancements integrate computer vision, ML, ethology, and neuroscience, exemplified by resources such as \textit{OpenBehavior} \cite{ref28} (https://edspace.american.edu/openbehavior) and \textit{TheBehaviourForum} (www.thebehaviourforum.org) which list community-developed tools and references specifically designed for social interaction analysis. Additional tools encompassing both individual and social behavior analysis are catalogued in \hyperref[tab:tool_list]{Supplementary Table 1}.

\textbf{In this methodological review, we provide an overview of current tools and methods for analyzing rodent social behavior, with particular focus on how machine learning is transforming this traditionally human observation-dependent field.}

% Table 1 - Full width using strip environment
\begin{table*}[ht]
\centering

\label{tab:human_annotation_limitations}
\scriptsize
\setlength{\tabcolsep}{4pt}
\renewcommand{\arraystretch}{1.12}

\begin{tabularx}{\textwidth}{L{3.8cm}*{12}{C{0.7cm}}}
\toprule
& \rotatebox{60}{Eco-HAB \cite{ref33}}
& \rotatebox{60}{LiveMouseTracker \cite{ref34}}
& \rotatebox{60}{3D-Tracker \cite{ref35}}
& \rotatebox{60}{DeepBehavior \cite{ref41}}
& \rotatebox{60}{Hong WF \cite{ref36}}
& \rotatebox{60}{SIPEC \cite{ref51}}
& \rotatebox{60}{3DDD SMT \cite{ref37}}
& \rotatebox{60}{AlphaTracker \cite{ref45}}
& \rotatebox{60}{DeepEthogram \cite{ref38}}
& \rotatebox{60}{Keypoint-MoSeq \cite{ref54}}
& \rotatebox{60}{SimBA \cite{ref56}}
& \rotatebox{60}{MARS \cite{ref46}}\\
\midrule
\multicolumn{13}{l}{\bfseries Addressed limitations of human annotation}\\
\gaprow

\textbf{Time \& scalability}      & 
\cmark&\cmark&\cmark&\cmark&\cmark&\cmark&\cmark&\cmark&\cmark&\cmark&\cmark&\cmark\\[6pt]
\defnrow{Manual annotation is time-consuming, often demanding two to three times the length of the video for accurate behavior annotation. The labor-intensive nature of manual work hampers broad generalization of the results, longer recordings and in-depth analysis..}
\gaprow
\gaprow

\textbf{Reproducibility}          & \cmark&\cmark&\cmark&\cmark&\cmark&\cmark&\cmark&\cmark&\cmark&\cmark&\cmark&\cmark\\[6pt]
\defnrow{Lack of standardization and high annotation variability between evaluators and sessions leads to inconsistencies in data, which can affect comparisons across studies and validation of results.}
\gaprow
\gaprow

\textbf{Temporal precision}          & \dash&\cmark&\cmark&\cmark&\dash&\cmark&\cmark&\cmark&\cmark&\cmark&\cmark&\cmark\\[6pt]
\defnrow{The low temporal precision of behavior time-stamp or recognition relative to their actual real occurrence impacts the alignment and correlation of behavior with high-resolution physiological data, possibly leading to misinterpretations of the timing and sequence of events.}
\gaprow
\gaprow

\textbf{Granularity}          & \dash&\dash&\dash&\dash&\dash&\dash&\dash&\cmark&\dash&\cmark&\dash&\dash\\[6pt]
\defnrow{The low degree of detail in classifying and observing behaviors may oversimplify or ignore subtle behavior variations, resulting in over-simplification or excessive categorization, failing to accurately reflect the true behavioral spectrum.}
\gaprow
\gaprow

\textbf{Anthropomorphism}          & \dash&\dash&\dash&\dash&\dash&\dash&\dash&\cmark&\dash&\cmark&\dash&\dash\\[6pt]
\defnrow{Characterization of behavior is defined by human language and interpretability, limiting our understanding of its biological basis.}
\gaprow

\midrule

\multicolumn{13}{l}{\bfseries Applications in Social Neuronal Activity Studies}\\
\gaprow

\textbf{Correlation with in vivo Neuronal Recording}      & \dash&\cmark&\cmark&\dash&\dash&\dash&\dash&\dash&\dash&\dash&\cmark&\cmark\\[6pt]
\gaprow

\textbf{Causality with real-time Neuronal Activity}          & \dash&\dash&\dash&\dash&\dash&\dash&\dash&\dash&\dash&\dash&\dash&\dash\\[6pt]

\bottomrule
\end{tabularx}
\caption{\textbf{Key limitations of human annotation of social behavior in rodent studies and how tools adresses these limitations and are applied to neuroscience studies.} The constraints include time and scalability issues, lack of reproducibility, limited temporal precision, coarse granularity, and anthropomorphic biases. The table also provides an overview of the tools' applications in studying neuronal activity during social behavior, including correlation with in vivo recordings and causality with real-time neuronal activity manipulation.}
\end{table*}

Unlike other recent reviews that discuss general progress of behavioral neuroscience through illustrative \textit{application} of computational tools \cite{ref23,ref24,ref25,ref26,ref29,ref30,ref31,ref32}, our work specifically addresses the unique challenges and advancements in the \textit{methods} studying rodent social behavior.

\begin{figure*}[!ht]
\centering
\includegraphics[width=0.9\textwidth]{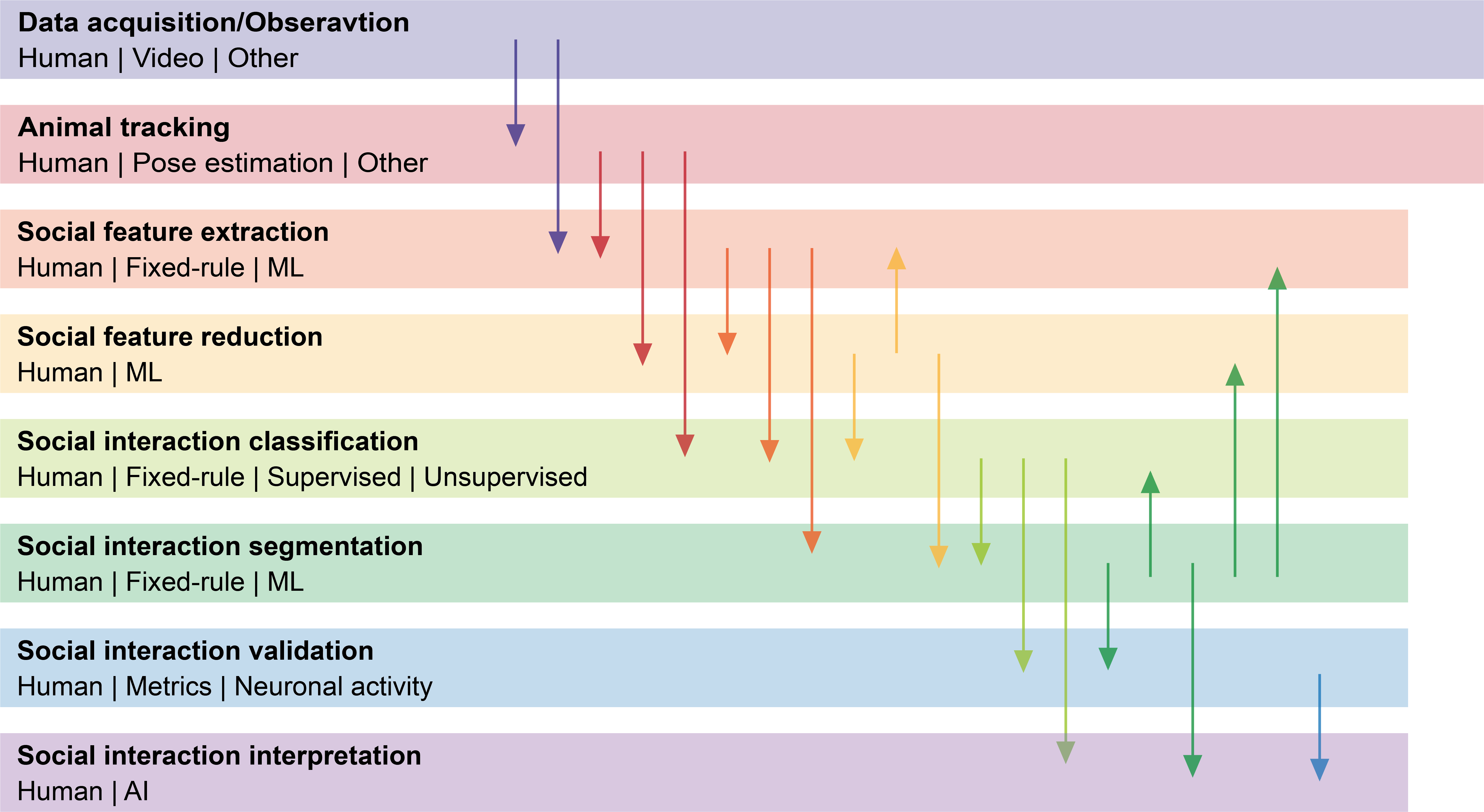}
\caption{\textbf{Overview of the typical analysis pipeline for studying social interaction in rodents.} The pipeline is decomposed into a series of steps: data acquisition, animal tracking, social feature extraction, social feature reduction, social interaction classification, social interaction segmentation, social interaction validation, and social interaction interpretation. For each step, the most established tools supporting it are referenced.}
\label{fig:pipeline}
\end{figure*}

Our review is structured around a pipeline encompassing data acquisition, animal tracking, social feature extraction and reduction, interaction classification, segmentation, validation, and interpretation \hyperref[fig:pipeline]{(Figure 1)}. Importantly, we clearly delineate each step to facilitate integration into cohesive analytical frameworks, emphasizing their contributions specifically within the context of social interaction analysis. Each pipeline component may vary in reliance on human input versus ML automation, and different tools can focus on various steps or be combined into a complete pipeline \hyperref[tab:tool_comparison]{(Table 2)}. We focus on tools proven effective in social scenarios, as single-animal methods often fail to scale to multiple individuals due to technical challenges (e.g., maintaining individual identities during occlusions), invalid assumptions, and sparse training data. Focus boxes throughout provide practical guidance on data acquisition strategies, performance evaluation metrics, and design considerations. We also introduce \textit{BANOS (Behavior Annotation Score)}, a comprehensive evaluation package addressing biases in both human and algorithmic analyses. Our goal is to equip researchers with the methodological knowledge and practical tools necessary to effectively study complex rodent social behaviors using contemporary machine learning approaches.

% Figure 1 - Full width

\section{Data acquisition / observation}

Data acquisition involves obtaining behavioral data using recording systems such as cameras, sensors, or direct human observation. Recorded data, especially videos, offer significant advantages over direct human observation by allowing consistent, reproducible, and iterative analysis. Beyond video recordings, other modalities have proven valuable for analyzing social interactions, including Radio Frequency IDentification (RFID) tracking and depth sensor-based 3D reconstructions. Acquiring data for social behavior studies involves specific considerations. Cameras must ensure all interacting animals remain visible throughout the recording, often requiring multiple viewpoints. RFID tracking is particularly helpful for maintaining individual animal identities during visual occlusions.

Most tools for social behavior analysis do not include data acquisition but instead rely on externally collected video data. Generic software can handle video recordings prior to more specialized analysis, but multimodal data acquisition (e.g., combining neuronal and video data) requires careful synchronization strategies, such as using an analog TTL (Time To Live) signal to align behavior and brain activity precisely. Ensuring consistent bandwidth and sampling rates is crucial to prevent signal drift over time. Hardware choices significantly affect recording quality and subsequent algorithm performance. For instance, global shutter cameras can substantially reduce motion-induced artifacts, enhancing animal identification and tracking accuracy.

Several tools support specialized data acquisition setups, including \textit{Eco-HAB} \cite{ref33}, \textit{LiveMouseTracker} \cite{ref34}, \textit{3Dtracker} \cite{ref35}, \textit{Hong workflow} \cite{ref36}, \textit{3DDD Social Mouse Tracker} \cite{ref37} and \textit{DeepEthogram} \cite{ref38}. Except for \textit{DeepEthogram} \cite{ref38}, these tools typically require specific equipment setups such as depth sensors or RFID tracking. This can reduce pipeline flexibility, although tools like \textit{3Dtracker} \cite{ref35} and \textit{3DDD Social Mouse Tracker} \cite{ref37} allow some customization through calibration.

Ensuring data quality and relevance at this stage is essential, as it influences all subsequent analysis steps. Additional considerations are discussed in \hyperref[box:data_acquisition]{Focus Box 1}.

\section{Animal tracking}

\begin{figure*}[!ht]
\centering
\includegraphics[width=0.9\textwidth]{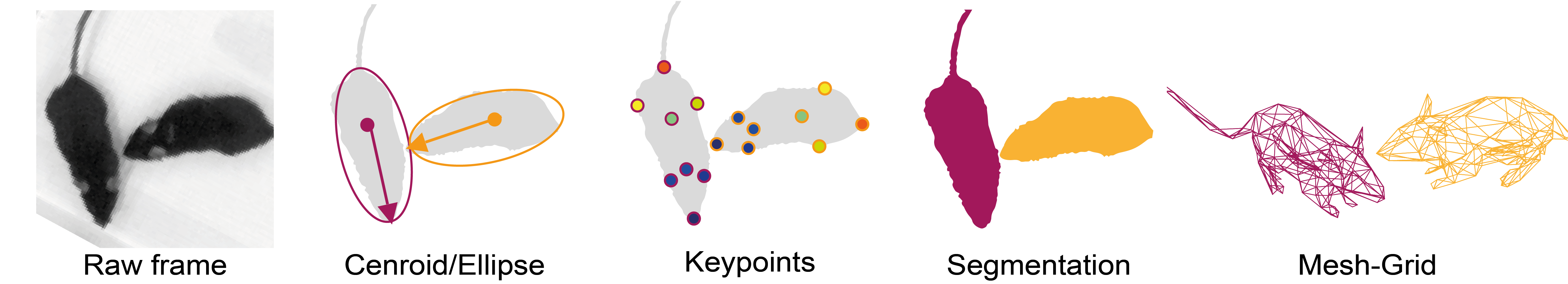}
\caption{\textbf{Illustration of common tracking strategies used to extract animal movement information from video recordings of rodents.} These include centroid/ellipse tracking, keypoint tracking, image segmentation, and mesh-grid tracking.}
\label{fig:tracking}
\end{figure*}

Raw behavioral data must be processed into a clear and structured format suitable for analyzing social interactions. This involves tasks like filtering background noise, tracking animals, correcting data errors, and normalizing scales. Depth sensing and RFID data typically require minimal preprocessing, whereas video data demand extensive refinement due to variability from factors like recording equipment, lighting conditions, and resolution. Among the different preprocessing tasks, animal tracking is particularly important to transform raw video into structured data for subsequent analysis steps. Several tracking strategies exist \hyperref[fig:tracking]{(Figure 2)}:

\begin{itemize}[leftmargin=*,
                nosep,           % removes the extra top & bottom glue
                itemsep=2pt,     % vertical space *between* items
                parsep=0pt,      % space after a paragraph inside an item
                label=\textbullet] % keep your usual bullet
\item \text{Centroid/Ellipse:} tracks the center (and sometimes the orientation) of animal bodies, providing x and y coordinates.
\item \text{Keypoints:} tracks specific body parts selected through pose estimation, providing x and y coordinates. For a review of body tracking, see  \cite{ref39}.
\item \text{Image Segmentation:} isolates animals from the background to generate masks.
\item \text{Mesh-grid:} uses intersecting lines to present spatial structure.
\end{itemize}

Identifying animals accurately, especially in multi-animal contexts, poses unique computational challenges. Tools supporting multiple animal keypoint tracking include \textit{DeepLabCut} \cite{ref40}, \textit{DeepBehavior} \cite{ref41}, \textit{SLEAP} \cite{ref42}, \textit{DANNCE} \cite{ref43}, \textit{SIPEC} \cite{ref44}, \textit{3DDD Social Mouse Tracker} \cite{ref37}, \textit{AlphaTracker} \cite{ref45} and \textit{MARS} \cite{ref46}. Tracking identity across multiple animals is supported by tools like \textit{LiveMouseTracker} \cite{ref34}, \textit{ToxTrac} \cite{ref47}, \textit{DeepLabCut} \cite{ref40}, \textit{idtracker.ai} \cite{ref48}, \textit{DeepBehavior} \cite{ref41}, \textit{TRex} \cite{ref49}, \textit{SLEAP} \cite{ref42}, \textit{DANNCE} \cite{ref43}, AnimalTA \cite{ref50}, \textit{Hong workflow} \cite{ref36}, \textit{SIPEC} \cite{ref51}, \textit{3DDD Social Mouse Tracker} \cite{ref37}, \textit{AlphaTracker} \cite{ref45} and \textit{MARS} \cite{ref46}. Among this list of tools tracking identity of multiple animals, \textit{MARS} \cite{ref46} supports only multiple animals with different fur (black vs white) and \textit{LiveMouseTracker} \cite{ref34} corrects the identity of animals through RFID tracking. Certain tools (e.g., \textit{DeepLabCut} \cite{ref40}, \textit{DeepBehavior} \cite{ref41}, \textit{Anipose} \cite{ref52}, \textit{TRex} \cite{ref49}, \textit{LiftPose3D} \cite{ref53} and \textit{SIPEC} \cite{ref44}) also support 3D tracking, while many others are limited to 2D. Some systems, such as \textit{DeepEthogram} \cite{ref38}, bypass animal tracking but are more sensitive to changes in recording conditions and less generalizable to different setups.

% Table 2 - Landscape page
\begin{landscape}
\begin{table}[p]
\centering
\label{tab:tool_comparison}
\scriptsize
\setlength{\tabcolsep}{2pt}
\renewcommand{\arraystretch}{1.1}

\begin{adjustbox}{width=\linewidth,center}
\begin{tabular}{p{2.5cm}p{3cm}*{21}{c}}
\toprule
& & \rotatebox{60}{\textbf{Eco-HAB} \cite{ref33}}
& \rotatebox{60}{\textbf{LiveMouseTracker} \cite{ref34}}
& \rotatebox{60}{\textbf{3DTracker} \cite{ref35}}
& \rotatebox{60}{\textbf{ToxTrac} \cite{ref47}}
& \rotatebox{60}{\textbf{DeepLabCut} \cite{ref40}}
& \rotatebox{60}{\textbf{idtracker.ai} \cite{ref48}}
& \rotatebox{60}{\textbf{DeepBehavior} \cite{ref41}}
& \rotatebox{60}{\textbf{TRex} \cite{ref49}}
& \rotatebox{60}{\textbf{Anipose} \cite{ref52}}
& \rotatebox{60}{\textbf{LiftPose3D} \cite{ref53}}
& \rotatebox{60}{\textbf{SLEAP} \cite{ref42}}
& \rotatebox{60}{\textbf{DANNCE} \cite{ref43}}
& \rotatebox{60}{\textbf{AnimalTA} \cite{ref50}}
& \rotatebox{60}{\textit{\textbf{Hong Workflow}} \cite{ref36}}
& \rotatebox{60}{\textbf{SIPEC} \cite{ref51}}
& \rotatebox{60}{\textbf{3DDD Social Mouse Tracker} \cite{ref37}}
& \rotatebox{60}{\textbf{AlphaTracker} \cite{ref45}}
& \rotatebox{60}{\textbf{DeepEthogram} \cite{ref38}}
& \rotatebox{60}{\textbf{Keypoint-MoSeq} \cite{ref54}}
& \rotatebox{60}{\textbf{SimBA} \cite{ref56}}
& \rotatebox{60}{\textbf{MARS} \cite{ref46}}\\
\midrule

% Data acquisition
\multicolumn{2}{l}{\textbf{Data acquisition}} &
  \cmark&\cmark&\cmark&\dash&\dash&\dash&\dash&\dash&\dash&\dash&\dash&\dash&\dash&\cmark&\dash&\cmark&\dash&\cmark&\dash&\dash&\dash\\
\gaprow

% Animal tracking
\multirow{6}{*}{\textbf{Animal tracking}} 
& \textbf{Video tracking} &
  \dash&\cmark&\dash&\cmark&\cmark&\cmark&\cmark&\cmark&\cmark&\cmark&\cmark&\cmark&\cmark&\cmark&\cmark&\cmark&\cmark&\dash&\dash&\dash&\cmark\\
& \hspace{0.5em}Multiple keypoints &
  \dash&\dash&\dash&\dash&\cmark&\dash&\cmark&\dash&\dash&\dash&\cmark&\cmark&\dash&\dash&\cmark&\cmark&\cmark&\dash&\dash&\dash&\cmark\\
& \hspace{0.5em}Multiple animal (a) &
  \dash&\cmark&\dash&\cmark&\cmark&\cmark&\cmark&\cmark&\dash&\dash&\cmark&\cmark&\cmark&\cmark&\cmark&\cmark&\cmark&\dash&\dash&\dash&(h)\\
& \hspace{0.5em}3D reconstruction keypoints &
  \dash&\dash&\dash&\dash&\cmark&\dash&\cmark&\dash&\cmark&\cmark&\dash&\dash&\dash&\dash&\cmark&\dash&\dash&\dash&\dash&\dash&\dash\\
& \textbf{3D depth sensing} &
  \dash&\cmark&\cmark&\dash&\dash&\dash&\dash&\dash&\dash&\dash&\dash&\dash&\dash&\cmark&\dash&\cmark&\dash&\dash&\dash&\dash&\dash\\
& \textbf{RFID multiple animal} &
  \cmark&\cmark&\dash&\dash&\dash&\dash&\dash&\dash&\dash&\dash&\dash&\dash&\dash&\dash&\dash&\dash&\dash&\dash&\dash&\dash&\dash\\
\gaprow

% Social feature extraction
\multirow{2}{*}{\parbox{2.5cm}{\textbf{Social feature\\extraction}}} 
& \textbf{Fixed-rule} &
  \cmark&\cmark&\cmark&\dash&\dash&\dash&\cmark&\dash&\dash&\dash&\dash&\dash&\dash&\cmark&\dash&\cmark&\cmark&\dash&\dash&\cmark&\cmark\\
& \textbf{ML} &
  \dash&\dash&\dash&\dash&\dash&\dash&\dash&\dash&\dash&\dash&\dash&\dash&\dash&\dash&\cmark&\dash&\dash&\cmark&\cmark&\dash&\dash\\
\gaprow

% Social interaction classification
\multirow{3}{*}{\parbox{2.5cm}{\textbf{Social interaction\\classification}}} 
& \textbf{Fixed-rule} &
  \dash&\cmark&\cmark&\dash&\dash&\dash&\cmark&\dash&\dash&\dash&\dash&\dash&\dash&\dash&\dash&\cmark&\dash&\dash&\dash&\dash&\dash\\
& \textbf{Supervised} &
  \dash&\dash&\dash&\dash&\dash&\dash&\dash&\dash&\dash&\dash&\dash&\dash&\dash&\cmark&\cmark&\dash&\dash&\cmark&\dash&\cmark&\cmark\\
& \textbf{Unsupervised} &
  \dash&\dash&\dash&\dash&\dash&\dash&\dash&\dash&\dash&\dash&\dash&\dash&\dash&\dash&\dash&\dash&\cmark&\cmark&\dash&\cmark&\dash\\
\gaprow

% Social interaction segmentation
\multirow{2}{*}{\parbox{2.5cm}{\textbf{Social interaction\\segmentation}}} 
& \textbf{Fixed-rule} &
  \dash&\cmark&\cmark&\dash&\dash&\dash&\cmark&\dash&\dash&\dash&\dash&\dash&\dash&\cmark&\cmark&\cmark&\cmark&\cmark&\dash&\cmark&\cmark\\
& \textbf{ML} &
  \dash&\dash&\dash&\dash&\dash&\dash&\dash&\dash&\dash&\dash&\dash&\dash&\dash&\dash&\dash&\dash&\dash&\cmark&\cmark&\dash&\dash\\
\gaprow

% Social interaction validation
\multicolumn{2}{l}{\textbf{Social interaction validation}} &
  \cmark&\cmark&\cmark&\dash&\dash&\dash&\cmark&\dash&\dash&\dash&\dash&\dash&\dash&\cmark&\cmark&\cmark&\cmark&\cmark&\cmark&\cmark&\cmark\\

\midrule

% Setup
\multirow{7}{*}{\textbf{Setup}} 

& Custom setup &
  \cmark&\cmark&\cmark&\dash&\dash&\dash&\dash&\dash&\dash&\dash&\dash&\cmark&\dash&\cmark&\dash&\dash&\dash&\dash&\dash&\dash&\dash\\
& Camera &
  \dash&\dash&\dash&\cmark&\cmark&\cmark&\cmark&\cmark&\cmark&\cmark&\cmark&\cmark&\cmark&\cmark&\cmark&\cmark&\cmark&\cmark&\cmark&\cmark&\cmark\\
& GPU support &
  \dash&\dash&\dash&\dash&\cmark&\cmark&\cmark&\cmark&\cmark&\cmark&\cmark&\cmark&\cmark&\dash&\cmark&\cmark&\cmark&\cmark&\cmark&\cmark&\cmark\\

& MathWorks MATLAB\textsuperscript{\textregistered} &
  \dash&\dash&\cmark&\dash&\dash&\dash&\cmark&\dash&\dash&\dash&\dash&\dash&\dash&\cmark&\dash&\dash&\dash&\dash&\dash&\dash&\dash\\
& Python\textsuperscript{TM} &
  \cmark&\cmark&\cmark&\dash&\cmark&\cmark&\cmark&\cmark&\cmark&\cmark&\cmark&\cmark&\cmark&\dash&\cmark&\cmark&\cmark&\cmark&\cmark&\cmark&\cmark\\
& GUI &
  \dash&\cmark&\cmark&\cmark&\cmark&\cmark&\dash&\cmark&\dash&\dash&\cmark&\dash&\cmark&\dash&\dash&\dash&\cmark&\cmark&\dash&\cmark&\cmark\\
& Other &
  \dash&(b)&(d)&(e)&\dash&\dash&\dash&\dash&\dash&\dash&\dash&\dash&(f)&\dash&\dash&\dash&\dash&\dash&\dash&\dash&\dash\\
  
\midrule

% Additional features
\multicolumn{2}{l}{\textbf{Real time usage}} &
  \dash&(c)&\dash&\dash&\cmark&\dash&\dash&\cmark&\dash&\cmark&\cmark&\dash&\dash&\dash&\dash&\dash&\dash&\dash&\dash&\dash&\dash\\
\multicolumn{2}{l}{\textbf{Generalizability to different setup}} &
  \dash&\dash&\cmark&\cmark&\cmark&\cmark&\cmark&\cmark&\cmark&\cmark&\cmark&\cmark&\cmark&\dash&\cmark&\cmark&\cmark&\cmark&\cmark&\cmark&\cmark\\
\multicolumn{2}{l}{\textbf{Online documentation and support}} &
  \cmark&\cmark&\cmark&\cmark&\cmark&\cmark&\cmark&\cmark&\cmark&\cmark&\cmark&\cmark&\cmark&\dash&\cmark&\cmark&\cmark&\cmark&\cmark&\cmark&\cmark\\
\multicolumn{2}{l}{\textbf{Social interaction benchmarking public data}} &
  \dash&\dash&\dash&\dash&\dash&\dash&\dash&\dash&\dash&\dash&\dash&\dash&\dash&\dash&\dash&\dash&\dash&\cmark&(g)&\cmark&\cmark\\

\midrule
% Popularity
\multirow{2}{*}{\textbf{Popularity}} & \textbf{Number of citations (i)} &
  19&58&26&330&1361&91&33&57&51&23&111&51&7&104&17&5&11&42&24&183&55\\
& \textbf{Date of publication} &
  2016&2019&2013&2018&2018&2019&2019&2021&2021&2021&2022&2021&2023&2015&2022&2022&2023&2021&2023&2020&2021\\
\bottomrule
\end{tabular}
\end{adjustbox}
\caption{\textbf{Comparison of commonly used tools for analyzing social interactions in rodents.} The table indicates which pipeline steps each tool supports, along with other key features such as hardware and software requirements, generalizability, documentation, benchmarking, and popularity metrics. (a) Only consider tools able to natively handle multiple animals (not 2 tracking iteration of 2 different color animals). (b) \textit{LiveMouseTracker} \cite{ref34} needs the installation of the Icy software for data acquisition. (c) Some social interaction can be analyzed in real time but more can be analyzed offline. Closed-loop applications are not supported. (d) \textit{3DTracker} \cite{ref35} needs installation of an executable software on \textit{Microsoft Windows\textsuperscript{\textregistered}} operating system for data acquisition. (e) \textit{ToxTrac} \cite{ref47} needs installation of an executable software on \textit{Microsoft Windows\textsuperscript{\textregistered}} operating system. (f) \textit{AnimalTA} \cite{ref50} needs installation of an executable software on \textit{Microsoft Windows\textsuperscript{\textregistered}} operating system. (g) \textit{Keypoint-MoSeq} \cite{ref54} used tracking of body points coordinates of 1 animal among the 2 presents in the dataset used for benchmark (\textit{CalMS21} dataset \cite{ref74}) (h) \textit{MARS} \cite{ref46} perform only tracking of multiple animals with different coat colors (black and white). (i) Number of citations as of May 2024.}
\end{table}
\end{landscape}

\onecolumn
% Figure 2 - Full width
\begin{focusbox}[breakable, enhanced, width=\textwidth]{Focus Box 1: Considerations on data acquisition and analysis}
\label{box:data_acquisition}
In rodent behavior analysis, labs often struggle to integrate old and new datasets. This task is complicated by the rapid evolution of analysis tools, which frequently introduce new data standards. While unanimous consensus on standards might constrain creativity, establishing reference practices could be a more balanced approach. Moving from traditional sparse information stored in lab notebooks to modern dataset suitable for ML requires standardization and a systematic data acquisition approach, emphasizing the quality of behavioral tracking data. Initiatives further underscore the importance of structured, FAIR-compliant repositories for storing and sharing large-scale behavioral data, facilitating reproducibility, integrative analysis, and cross-laboratory collaboration \cite{ref99}. Investing in proper recording setups is also essential. Specifically, standardization can be enhanced by adopting uniform protocols for data capture, such as using consistent recording equipment and settings across all studies to ensure comparability. Additionally, employing common data formats and annotation guidelines can facilitate data sharing and analysis. A systematic approach might involve implementing structured data collection schedules, rigorous training for personnel on data collection procedures, and using automated systems to reduce human error. These measures ensure that the data collected is not only consistent across different times and settings but also reliably captures the necessary details for in-depth behavioral analysis.

\vspace{0.3cm}

Across laboratories, lab setups range from basic to enriched environments \cite{ref34,ref100,ref101}, each presenting unique analytical challenges. While enriched environments can offer deeper insights, they also complicate data analysis. Sophisticated tracking and isolation techniques are necessary due to the variety of interactions and background elements. The complexity of behaviors demands advanced feature extraction and data analysis methods to handle non-linear relationships and increased variability. Interpreting results and ensuring reproducibility also require careful consideration of environmental influences and detailed documentation of setups. However, advancements like keypoint tracking, image segmentation and depth sensing are driving data standardization in the field. Keypoint tracking consistently identifies and tracks specific anatomical points to generate spatial coordinates. Image segmentation isolates subjects from their backgrounds to generate a mask. Depth sensing measures the distance between the sensor and the subject to generate three-dimensional spatial data. Each of these techniques has the desirable side effect of removing noise factors such as brightness or background from the data, simplifying the downstream steps. Collectively, these technologies facilitate consistent and accurate tracking of subjects across various research setups.

\vspace{0.3cm}

For researchers interested in recording mice during social interaction using keypoint tracking, we suggest:

\begin{itemize}[leftmargin=*,
                nosep,           % removes the extra top & bottom glue
                itemsep=2pt,     % vertical space *between* items
                parsep=0pt,      % space after a paragraph inside an item
                label=\textbullet] % keep your usual bullet
\item Standardizing environmental conditions (e.g., lighting, background) to optimize video quality for ML tools;
\item Employing an adequately sized space, such as an open arena measuring 30x30 cm, provides an effective balance between compactness and allowing animals to interact voluntarily without being forced. Use appropriate arena sizes that allow for a full range of social behaviors while keeping animals in frame;
\item Considering the use of multiple camera angles to capture all aspects of social interactions. Setup can be optionally equipped with mirrors for side views to aid reconstruction and enable 3D modeling using a single camera \hyperref[fig:3d_acquisition]{(see Supplementary Figure 2)};
\item Ensuring consistent animal identification (e.g., through distinct markings or RFID tags);
\item Recording at a sufficient rate. For example, 25 fps (frame per second) is a good default for most social interaction analysis. However, recording details of movement during event like attack or of body parts like paw or whisker may require a higher acquisition rate. Collecting high frame rate videos (e.g., 60 fps) is recommended to capture rapid social interactions;
\item Focusing on keypoint tracking of selected body joints. The choice of relevant points depends on the specific application and research questions. See examples of key points tracking configuration in \hyperref[fig:keypoints]{Figure 3}. However, to our knowledge, this selection remains largely arbitrary and intuition-based. While tools like \textit{A-SOiD} \cite{ref77} have shown effective classification resilience when reducing keypoints set, systematic studies on how different configurations impact behavior discovery in unsupervised settings are still lacking. Generally, more keypoints may enable detection of subtle behaviors, though strategic selection of the most informative points could prove equally effective;
\item Including likelihood estimates for each point with tracking, which quantify the tracking algorithm's confidence in the accuracy of the identified keypoints;
\item Providing essential metadata, including temporal (fps) and spatial resolution (pixel to cm ratio);
\item Exporting data in a human-readable, machine-parsable format (e.g. .csv, .tsv or .json);
\item Storing multi-animal tracking data in separate files per subject to simplify the analysis of large groups;
\item Collecting sufficient data for training supervised ML models, typically several hours of diverse social interactions.
\end{itemize}

\vspace{0.3cm}

% Figure 3 - Full width
\begin{figure}[H]
\centering
\includegraphics[width=0.9\textwidth]{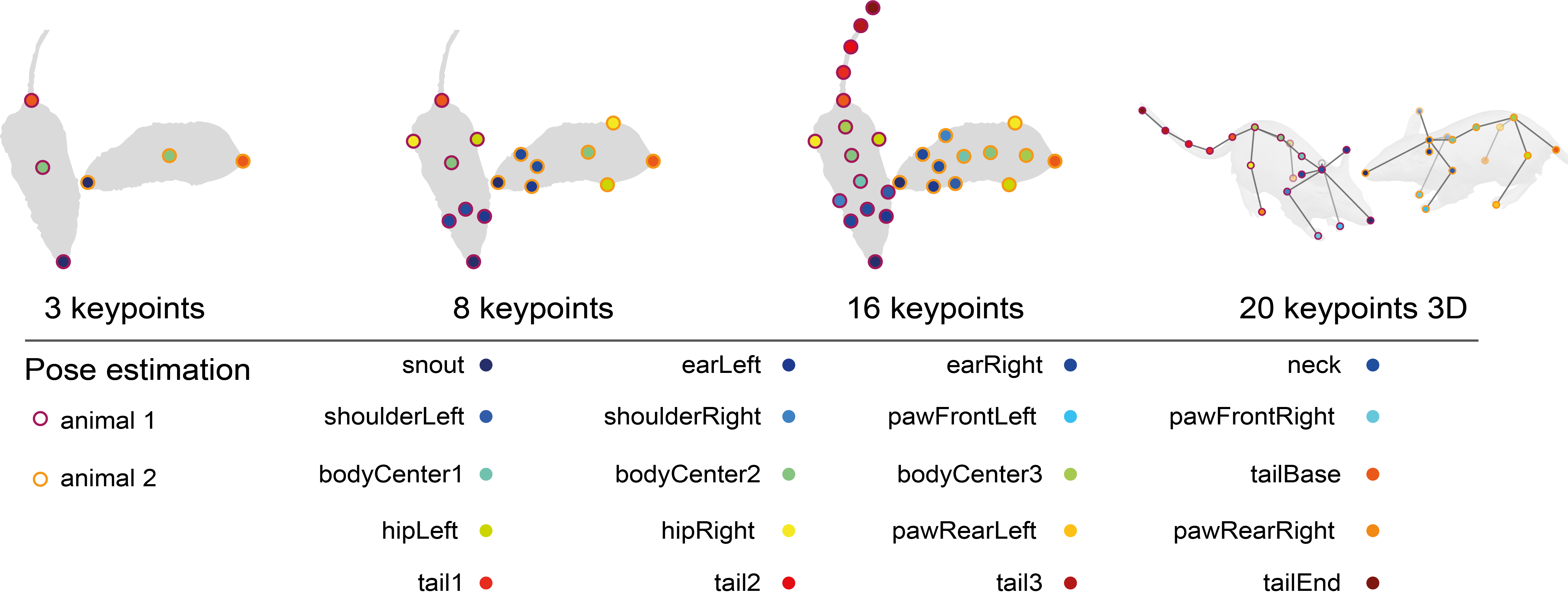}
\caption{\textbf{Anatomical keypoint configurations commonly used for pose estimation in rodents.} The choice of keypoints depends on the specific research application and questions. More keypoints can potentially help detect more subtle behaviors.}
\label{fig:keypoints}
\end{figure}

The alignment between experimental setups and data must accommodate both manual analysis by humans and automated processing by computers. For example, this involves formatting data so that it is readily accessible in common software like \textit{Microsoft Excel\textsuperscript{\textregistered}} for human use, and structured appropriately for computational tools used in more advanced analyses. The practical value of data is greatly enhanced when it is easily usable in both these contexts, ensuring it meets the diverse needs of the research community. 
For researchers interested in implementing machine learning tools for social behavior analysis, careful consideration of tool selection and implementation is crucial. The choice of tools should be guided primarily by the specific research questions being addressed and the types of social behaviors under investigation. Different tools may excel at capturing various aspects of social interactions, so aligning the tool capabilities with the study's focus is essential. Once potential tools have been identified, it is imperative to validate their performance on a subset of the research data before proceeding with full-scale implementation. This validation step helps ensure that the chosen tools are suitable for the particular experimental setup and can accurately detect and classify the behaviors of interest. 

Additionally, researchers should consider employing ensemble approaches, which involve combining multiple tools to leverage their respective strengths. This strategy can often lead to more robust and comprehensive analyses, as different tools may excel at detecting different aspects of social behavior.

\vspace{0.3cm}

In data analysis, implementing rigorous cross-validation procedures is vital to ensure model generalizability. This process helps verify that the machine learning models can accurately analyze new, unseen data and are not overfitting to the specific characteristics of the training dataset. Furthermore, researchers should not solely rely on statistical significance when interpreting machine-detected behaviors. It is equally important to consider the ethological relevance of these behaviors, ensuring that the identified patterns align with known or plausible biological phenomena and contribute meaningful insights to the field of study.

\vspace{0.3cm}

When it comes to reporting results, researchers should provide a clear and detailed description of all steps in their machine learning pipeline, including the specific hyperparameters used and the training procedures employed. This level of detail allows other researchers to understand the exact methodology and potentially replicate the study. It is crucial to report performance metrics based on held-out test sets rather than just the training data, as this gives a more accurate representation of the model's true predictive capabilities. Additionally, sharing both the code used for analysis and the trained models themselves greatly enhances the reproducibility of the study.
By following these guidelines, researchers can improve the quality, consistency, and reproducibility of their rodent behavior analysis studies, while also facilitating the integration of traditional and modern ML-based approaches.
\end{focusbox}
\twocolumn

Keypoint tracking introduces specific challenges affecting downstream analysis. Keypoint jittering, rapid fluctuations in tracked point positions due to estimation noise, can create false patterns that algorithms might interpret as meaningful behaviors. This is particularly problematic when algorithms are sensitive to high-frequency components in movement data.

\textit{Keypoint-MoSeq} \cite{ref54} and \textit{'neuroinformatics-unit/movement'} \cite{ref55} \textit{Python}$^{\text{TM}}$ toolbox have implemented preprocessing solutions to address this issue through noise filtering and probabilistic modeling accounting for keypoint uncertainty. Other common issues include missing keypoints due to occlusions (requiring interpolation strategies), quantization errors from discretization of coordinates, and misidentified body parts during complex interactions. These tracking artifacts can propagate through the analysis pipeline and significantly affect behavior annotation results.

Multi-animal tracking presents additional complexities, including frequent visual occlusions during close interactions, maintaining consistent animal identities, and higher computational demands increasing with the number of animals, particularly for live (closed-loop) tracking. Offline, tools like \textit{idtracker.ai} \cite{ref48} are specifically designed to maintain individual identities in challenging scenarios.

Automated tracking tools significantly enhance standardization and comparability of behavioral observations. They enable processing large datasets consistently, facilitating larger samples and longer observations than manual scoring. They also provide precise quantitative measurements of postures and movements, offering higher temporal resolution and reducing variability from human observers. This precise tracking ensures reliable and structured data \hyperref[box:data_acquisition]{(see Focus Box 1)}
 for subsequent analysis stages, forming a critical foundation for a behavioral analysis pipeline.

\section{Social feature extraction}

Social feature extraction involves identifying and quantifying key aspects of behavioral data that specifically represent interactions between animals. Unlike individual behavior analysis, social interactions require capturing relational dynamics, such as inter-animal distances, relative orientations, synchronized movements, and approach-avoidance patterns. Traditionally, humans intuitively assign behavioral scores, analogous to score sheets in animal welfare assessments, where observed behaviors are translated into quantifiable metrics for comprehensive evaluation. Social feature extraction is not always implemented as an explicit step. Some tools for example perform feature extraction, classification, and segmentation simultaneously, while other workflows bypass explicit feature extraction by directly processing preprocessed data for classification or segmentation.

Rule-based approaches systematically extract features using predefined rules on distances, angles, and body movements. Tools employing this strategy include \textit{Eco-HAB} \cite{ref33}, \textit{LiveMouseTracker} \cite{ref34}, \textit{3Dtracker} \cite{ref35}, \textit{DeepBehavior} \cite{ref41}, \textit{Hong workflow} \cite{ref36}, \textit{3DDD Social Mouse Tracker} \cite{ref37}, \textit{AlphaTracker} \cite{ref45}, \textit{SimBA} \cite{ref56} and \textit{MARS} \cite{ref46}. Despite advances in automatic extraction of preselected features, comprehensively defining social features to fully capture interaction complexity remains challenging.

Machine learning approaches have emerged to automatically discover complex behavioral patterns, enabling nuanced feature extraction beyond human intuition. Several tools originally designed for individual behavior analysis have been tested in social contexts. These include \textit{MoSeq} \cite{ref57}, \textit{B-SoiD} \cite{ref58}, \textit{VAME} \cite{ref59}, and \textit{MotionMapper} \cite{ref60}, which use machine learning to generate latent representations of behaviors. Among these, \textit{SIPEC} \cite{ref51} and \textit{DeepEthogram} \cite{ref38} utilize convolutional neural networks directly from video images, while \textit{Keypoint-MoSeq} \cite{ref54} applies \textit{Autoregressive Hidden Markov Models} for simultaneous feature extraction, classification, and segmentation.

Specialized social behavior tools, not yet referenced on \textit{OpenBehavior}, have been developed to address the unique complexities of social interactions more directly, including \textit{DeepOF} \cite{ref61}, \textit{SBeA} \cite{ref62}, \textit{LISBET} \cite{ref63} and \textit{BAMS} \cite{ref64}. \textit{DeepOF} \cite{ref61} and \textit{SBeA} \cite{ref62} leverage different artificial neural networks to extract relevant features, while \textit{BAMS} \cite{ref64} employs temporal convolutional networks operating at multiple timescales to capture rapid and slow behavioral dynamics simultaneously. \textit{LISBET} \cite{ref63} introduces a novel transformer-based approach via self-supervised learning, creating artificially altered scenarios to help the model distinguish between genuine and manipulated social interaction data without human labels. By explicitly focusing on relational aspects, \textit{LISBET} \cite{ref63} is biased toward extracting features related to social behaviors rather than individual actions, though it is currently limited to dyadic or multi-dyadic interactions rather than collective group behavior analysis.

Architectural considerations for temporal and social scales are crucial for effective feature extraction. Model architectures differ fundamentally in capturing temporal dependencies from milliseconds to minutes through choices like window sizes, recurrence and attention. Transformer-based models like \textit{LISBET} \cite{ref63} excel at discovering relationships within bounded timeframes through attention mechanisms that directly link distant timepoints, making them effective for complex behavioral dependencies. Temporal convolutional networks like \textit{BAMS} \cite{ref64} create multiple processing streams that analyze data at different temporal resolutions, making them suitable for comparing patterns across widely different timescales that would be computationally prohibitive for transformers. 

These architectural choices operate independently from classification approaches, both extraction and classification can be supervised or unsupervised, enabling hybrid approaches particularly efficient when labeled data is scarce.

Beyond temporal considerations, behaviors span multiple social scales: individual actions (e.g., rearing, grooming), dyadic interactions involving coordinated responses between two animals (e.g., fighting), and collective group behaviors from multiple individuals (e.g., coordinated hunting). Many tools are optimized for individual behaviors, with no comprehensive strategy tailored to social behaviors. While some can detect social behaviors by observing one animal's patterns during interaction (assuming that certain social behaviors have distinctive individual signatures like in fighting or mating), this approach could in theory be limited to interactions with clear individual behavioral markers. For instance, distinguishing between fighting and seizures might be challenging when considering only individual actions rather than dyadic interactions.

Similarly, tools optimized for extracting features related to dyadic interactions may be limited in fully capturing collective behaviors. To our knowledge, no tool has yet implemented a strategy to specifically handle collective group behavior beyond multi-dyadic interactions.

Emerging approaches show promise for advancing social behavior analysis. Foundation models like \textit{LlaMA} \cite{ref65} and \textit{LlaVA} \cite{ref66} contain rich, broadly applicable representations that could potentially be adapted for behavioral analysis through fine-tuning or transfer learning approaches. These models might recognize subtle behavioral patterns or contextual cues that domain-specific models miss, though their application to behavioral neuroscience remains largely unexplored.

Model interpretability represents a crucial frontier offering benefits: trust in systems and insights into behavioral organization. Tools like \textit{SHapley Additive exPlanations (SHAP)} audits identify which features most influence model decisions \cite{ref67}, potentially revealing previously unrecognized behavioral components. More advanced "AI biology" approaches \cite{ref68} enable researchers to trace computational circuits within models and understand causal relationships, using AI not just for annotation but as a discovery mechanism for understanding behavioral principles, revealing how models decompose complex social behaviors into component features.

\section{Social feature reduction}

This step focuses on reducing the dimensionality of the data, often by dropping or merging features.

It should be noted that dimensionality reduction is often a common secondary effect of the feature extraction phase. However, the feature space might still be too large for the effective application of the subsequent steps of the pipeline.

In social behavior analysis, feature reduction is generally performed using techniques such as \textit{Principal Component Analysis (PCA)} \cite{ref69}, \textit{t-Distributed Stochastic Neighbor Embedding (t-SNE)} \cite{ref70}, and \textit{Uniform Manifold Approximation and Projection (UMAP)} \cite{ref71}, even though these last two are often considered as data visualization techniques, due to their stochastic nature. To date, we are not aware of any dimensionality reduction techniques specifically engineered for processing social behavior data.

Recently, \textit{CEBRA} \cite{ref72} and \textit{MARBLE} \cite{ref73} have been proposed as a novel approach in the field of neuroscience for dimensionality reduction. Testing \textit{CEBRA} \cite{ref72} conditioned on time and \textit{MARBLE} \cite{ref73} with behavioral data in social configurations or integrating them with other tools may potentially enhance analysis visualization.

\section{Social interaction classification}

Social interaction classification involves categorizing extracted behavioral features into specific, meaningful labels. Human annotators typically perform this task intuitively, relying on subjective interpretations influenced by personal experience or implicit internal models. However, such subjectivity creates challenges, including inconsistencies in defining behavior categories, determining appropriate granularity, and understanding behavioral interrelationships. Accurately classifying social interactions is further complicated by simultaneous behaviors of multiple animals and context-dependent interactions, where subtle cues like intensity or preceding behaviors often determine whether an interaction is classified as playful or aggressive.

\subsection{Heuristic and Rule-Based Classification}

Heuristic methods enhance reproducibility and temporal precision by using explicit definitions based on measurable social features. For example, an "approach" behavior might be defined by specific proximity and speed thresholds \cite{ref74}. Tools employing this approach include \textit{LiveMouseTracker} \cite{ref34}, \textit{3Dtracker} \cite{ref35}, \textit{DeepBehavior} \cite{ref41} and \textit{3DDD Social Mouse Tracker} \cite{ref37}.

Recent advances in natural language processing have enabled large language models like \textit{ChatGPT} \cite{ref75} and interfaces like \textit{AmadeusGPT} \cite{ref76} to classify behaviors based on user-provided heuristic definitions through structured prompts. While these methods provide interpretable and reproducible classifications, they may fail to capture complex, subtle behaviors that require more nuanced pattern recognition.

\subsection{Supervised Learning Classification}

Supervised approaches address heuristic limitations by training models on human-annotated datasets to recognize patterns from labeled examples. Tools such as \textit{Hong workflow} \cite{ref36}, \textit{SIPEC} \cite{ref51}, \textit{DeepEthogram} \cite{ref38}, \textit{SimBA} \cite{ref56}, and \textit{MARS} \cite{ref46} employ this principle, offering frameworks that promote reproducible annotations. Once trained, these models can consistently replicate human-like annotations, which is invaluable for standardizing behavioral analysis across different studies. Supervised learning provides transparency by revealing influential features in decision-making but requires careful handling of challenges like overfitting, underfitting, data leakage, and training data quality.

Emerging hybrid methods, while not yet referenced through platforms like \textit{OpenBehavior}, combine supervised and unsupervised approaches to reduce annotation burden. \textit{A-SOiD} \cite{ref77} (and \textit{JABS} \cite{ref78} but not yet for social behavior) employs active learning to iteratively refine models using minimal labeled data, strategically querying edge-case examples to improve decision boundaries while mitigating over-representation of dominant behavior classes, enhancing efficiency and accuracy. The system includes an unsupervised discovery module that identifies and proposes sub-classes within existing labels for researcher validation. \textit{TREBA} \cite{ref79} offers a complementary approach through "task programming," where domain experts define interpretable behavioral metrics (e.g., inter-mouse distance, facing angle, speed) that guide self-supervised learning. This allows experts to encode structured knowledge once rather than perform repetitive labeling, achieving significant reductions in required annotated data while maintaining performance.

Challenges and limitations of supervised learning include dependency on rare, high-quality labeled data, with model efficacy heavily influenced by training data quality, size, and diversity. Practitioners must carefully handle overfitting, underfitting, and data leakage during training and evaluation. Critical decisions regarding input data selection (precomputed features, raw videos, tracking data) and hyperparameter calibration significantly impact performance. Additionally, supervised learning inherits human annotation challenges, including subjective category definitions and inter-annotator variability that can introduce inconsistencies across datasets.

\subsection{Unsupervised Learning Classification}

Clustering-based methods identify patterns without predefined labels using approaches such as \textit{hierarchical clustering} \cite{ref80}, \textit{k-means} \cite{ref81}, or \textit{DBSCAN} \cite{ref82}. Tools like \textit{3DDD Social Mouse Tracker} \cite{ref37}, \textit{AlphaTracker} \cite{ref45}, and \textit{Keypoint-MoSeq} \cite{ref54} apply unsupervised algorithms to classify and segment behaviors, enabling exploration beyond human-defined categories. \textit{Keypoint-MoSeq} \cite{ref54} notably uses \textit{Autoregressive Hidden Markov Models (AR-HMM)} to perform feature extraction, classification, and segmentation simultaneously.

Granularity management represent key challenges in unsupervised approaches. Behavioral granularity determines how finely behaviors are categorized in final output, whether systems identify "attack" as a single category or distinguish between "lunge," "bite," and "chase" as separate behaviors. Methods for handling output granularity vary significantly between tools. \textit{Keypoint-MoSeq} \cite{ref54} addresses this through manual control of the "stickiness" parameter, allowing researchers to tune temporal scales from fine-grained movements to broader behavioral states, though this requires user expertise and iterative testing. More sophisticated approaches include Bergman et al.'s multi-scale approach revealing hierarchical behavioral structure of fly behavior \cite{ref83} and \textit{LISBET}'s automatic multi-scale approach \cite{ref63}, which fits multiple Hidden Markov Models with different state numbers, then clusters similar motifs hierarchically to identify representative prototypes without pre-specifying category numbers. This multi-scale clustering approach developed in \textit{LISBET} \cite{ref63} could potentially be leveraged by other tools like \textit{Keypoint-MoSeq} \cite{ref54}, or even serve as a framework for integrating outputs from multiple annotation systems while avoiding redundancy.

Limitations include interpretability challenges due to absent predefined labels, requiring careful characterization of discovered categories (also called events, syllables, or motifs). Many algorithms require a priori specification of cluster numbers, potentially leading to over- or under-categorization, while advanced algorithms often demand substantial computational resources. Despite these challenges, unsupervised methods excel at systematically analyzing large datasets and capturing subtle behavioral variations that might escape human observation.

\subsection{Classification paradigms and considerations}

Most methods formulate social behavior annotation as multi-class problems, assuming mutually exclusive behaviors at each timepoint using one-hot encoding. However, some tools apply classification models to individual behavior classes, creating overlapping labels where multiple behaviors can occur simultaneously. While overlap reflects the complex, layered nature of social interactions \cite{ref24}, exclusivity may be desirable for certain behaviors. Strategies for handling overlaps include heuristic splitting by dividing behaviors temporally, assigning priority based on appearance order, or using predefined hierarchies. The choice between these approaches depends on the specific research questions and the nature of the behaviors being studied.

\section{Social interaction segmentation}

Humans naturally segment continuous behavior into meaningful events based on intuitive understanding, such as identifying the start and end of aggressive interactions. Similarly, behavioral segmentation in data analysis involves explicitly marking the beginnings and ends of social interaction episodes, distinguishing it from classification, which labels individual data points.

Behavioral segmentation can occur at different pipeline stages:

\begin{itemize}[leftmargin=*,
                nosep,           % removes the extra top & bottom glue
                itemsep=2pt,     % vertical space *between* items
                parsep=0pt,      % space after a paragraph inside an item
                label=\textbullet]
\item \text{Early-stage segmentation:} Defines social event boundaries prior to feature extraction and classification of the full segment
\item \text{Post-classification segmentation:} Identifies segment boundaries after classification, further refining the contextual understanding of each categorized event.
\end{itemize}

Behavioral segmentation can be conducted using heuristic or deterministic methods by a variety of tools, including \textit{LiveMouseTracker} \cite{ref34}, \textit{3Dtracker} \cite{ref35}, \textit{DeepBehavior} \cite{ref41}, \textit{Hong workflow} \cite{ref36}, \textit{SIPEC} \cite{ref51}, \textit{3DDD Social Mouse Tracker} \cite{ref37}, \textit{AlphaTracker} \cite{ref45}, \textit{DeepEthogram} \cite{ref38}, \textit{Keypoint-MoSeq} \cite{ref54}, \textit{SimBA} \cite{ref56}, and \textit{MARS} \cite{ref46}. For instance, \textit{SIPEC} \cite{ref51} restricts social events to periods of close proximity, such as a maximum distance threshold of 5 cm between animals \cite{ref17,ref84}.

However, defining start and end points can be arbitrary. This can result in data that may miss subtle but important aspects of social behavior, such as the approach phase before contact. Moreover, the continuity of segmented behaviors often does not align perfectly with human annotation, particularly when using methods that classify data frame by frame, where minor misclassifications (e.g., a single incorrect frame within a segment) disrupt continuity. This is common and can occur for various reasons like model inconsistency or tracking artifact, even of 1 single body part during 1 frame.

To improve behavioral segmentation fidelity, several strategies have been employed:

\begin{itemize}[leftmargin=*,
                nosep,           % removes the extra top & bottom glue
                itemsep=2pt,     % vertical space *between* items
                parsep=0pt,      % space after a paragraph inside an item
                label=\textbullet]
\item \text{Merging Rules:} combine contiguous segments separated by brief interruptions to maintain behavioral integrity.
\item \text{Duration Constraints:} Specify minimum and maximum segment durations to avoid overly fragmented or excessively lengthy segments.
\item \text{Data Smoothing:} Apply filtering techniques to reduce artifacts from tracking or classification errors.
\end{itemize}

More sophisticated machine learning algorithms, such as those used by \textit{AlphaTracker} \cite{ref45} or \textit{Keypoint-MoSeq} \cite{ref54}, employ techniques like \textit{Autoregressive Hidden Markov Models (AR-HMM)} to maintain relationships between consecutive timepoints during classification. This approach maximizes the creation of coherent segments that reflect the natural flow of behavior as closely as possible.

\onecolumn
\begin{focusbox}[breakable, enhanced, width=\textwidth]{Focus Box 2: Evaluating Classification Performance}
\label{box:banos}
Traditional metrics like F1 scores are used to evaluate how well a model replicates human annotations in rodent behavior studies (usually for supervised classification but also to compare annotation using other approaches like unsupervised classification). Specifically, these metrics are used on a frame-by-frame basis to compare the original annotation by a human (considered here as ground truth) with the predicted annotation by machine. The F1 score evaluates prediction performance by combining two metrics: precision, which measures the accuracy of positive predictions as the ratio of true positives to all predicted positives, and recall, which measures the ability to identify all actual positives as the ratio of true positives to actual positives. However, these metrics often do not suffice, as they do not capture the nature of rodent behaviors, which are more accurately represented as segments rather than discrete frames. This discrepancy results in differing objectives and understandings between biologists and machine learning engineers. Biologists aim for a multidimensional perspective to grasp various behavior prediction aspects, while machine learning engineers focus on optimizing algorithms with a single, overarching metric, potentially misaligning with biologists' expectations.

To bridge the gap between different objectives, we propose to use the following set of metrics to evaluate the quality of computer-assisted annotation \hyperref[fig:banos]{(see Figure 4)}:

\begin{itemize}[leftmargin=*,
                nosep,           % removes the extra top & bottom glue
                itemsep=2pt,     % vertical space *between* items
                parsep=0pt,      % space after a paragraph inside an item
                label=\textbullet]
\item \textbf{Detection Accuracy.} This measures the accuracy of identifying behavioral bouts using precision and recall to calculate the F1 score of the behavioral segment instead of the individual frames;
\item \textbf{Segment Overlap.} It assesses the temporal overlap of each annotated behavioral bout predicted with the ground truth, using temporal intersection over Union (tIoU), a measure of overlap between labels. This component could be useful for machine learning engineers, as offers a single comprehensive metric reflecting different prediction dimensions;
\item \textbf{Temporal Precision.} It evaluates the performance in predicting the start and end times of each behavior bout;
\item \textbf{Intra-bout Continuity.} It determines the continuity within each bout by quantifying the frequency of annotation changes, thus ensuring the consistency of behavioral annotations.
\end{itemize}

We collected these four metrics in an analysis package, the \textit{Behavior Annotation Scores (BANOS)}, featuring \textit{Python}$^{\text{TM}}$ and \textit{MATLAB}$^{\text{\textregistered}}$ implementations (https://github.com/BelloneLab/BANOS). We propose these metrics to provide biologists with detailed analysis tools aligned with their intuitive expectations and machine learning engineers with a better description of the requirements of this field of research, bridging disciplines and fostering understanding of evaluation criteria.

% Figure 4 - Full width
\begin{figure}[H]
\centering
\includegraphics[width=0.7\textwidth]{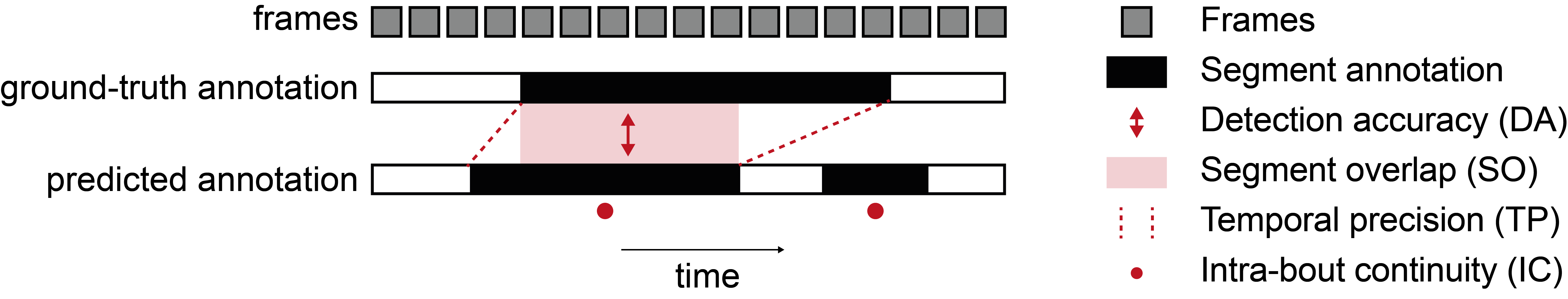}
\caption{\textbf{The \textit{Behavior Annotation Scores (BANOS)} metrics for evaluating the quality of computer-assisted behavioral annotations compared to human annotations.} \textit{BANOS} includes four key metrics: Detection Accuracy, which measures the precision and recall of behavioral segment predicted, using F1 score for segments instead of individual frames; Segment Overlap assesses the temporal Intersection over Union (tIoU) between predicted and ground truth annotated bouts; Temporal Precision evaluates the performance of start and end times predictions for each behavior segment; Intra-bout Continuity determines the consistency within bouts by quantifying annotation change frequency. \textit{BANOS} is available as a \textit{Python}$^{\text{TM}}$ and \textit{MathWorks MATLAB}$^{\text{\textregistered}}$ package at https://github.com/BelloneLab/BANOS.}
\label{fig:banos}
\end{figure}

\textit{BANOS} offers a different approach than other bout-level methods like those used in \textit{JABS} \cite{ref78}. While \textit{JABS} \cite{ref78}  employs Intersection over Union (IoU) with thresholds commonly used in computer vision (e.g., 0.3, 0.5, or 0.7) to evaluate detection performance, \textit{BANOS} avoids reliance on arbitrary thresholds by decomposing performance into four complementary metrics. \textit{JABS} \cite{ref78}  applies IoU thresholding to determine true positives, then calculates standard Precision, Recall, and F1 scores, typically presented as a function of IoU threshold. In contrast, \textit{BANOS} provides a multidimensional evaluation that captures different aspects of annotation quality (detection success, temporal alignment (using IoU), start/end precision, and internal consistency) without requiring subjective cutoff values. These metrics include and extend further the \textit{JABS} \cite{ref78}  approach to more nuanced diagnostic information that helps identify why a classifier might be failing at the ethological meaning level (e.g., late starts, fragmented bouts, or jittery predictions). For teams introducing new annotation models or comparing systems across labs, using both approaches can be valuable: \textit{JABS/BANOS-IoU} for standardized comparison with the ML community and \textit{BANOS} for deeper ethological insights.
In summary, the integration of computational tools into rodent behavior studies emphasizes the importance of data quality, an awareness of the limitations in human annotations, and the adoption of comprehensive evaluation metrics like those implemented in \textit{BANOS}. While recommending a threshold to consider a score as good for each metric would be arbitrary, we computed \textit{BANOS} on the \textit{CalMS21} dataset \cite{ref74} to give an idea of the scores that are expected to be obtained between 2 human annotators \hyperref[tab:banos_comparison]{(see TABLE 3)}. Interestingly, as already observed in other dataset \cite{ref46}, it seems that while human annotators show good agreement detecting presence or not of a behavior, they fail in agreeing on temporal precision of the behavior. 
\end{focusbox}
\twocolumn

\section{Social interaction validation}

Validating behavioral annotations involves ensuring that automatically annotated social interactions accurately reflect observed behaviors, a critical step before conducting deeper scientific analyses. Human validation involves visually confirming behavior labels frame-by-frame but remains inherently subjective and challenging to quantify. Most tools, including \textit{Eco-HAB} \cite{ref33}, \textit{LiveMouseTracker} \cite{ref34}, \textit{3Dtracker} \cite{ref35}, \textit{SIPEC} \cite{ref51}, \textit{AlphaTracker} \cite{ref45}, \textit{DeepEthogram} \cite{ref38}, \textit{SimBA} \cite{ref56}, and \textit{MARS} \cite{ref46}, facilitate such validation through human review interfaces.

Computational validation provides more objective assessment by comparing predicted annotations to ground truth labels using standard machine learning metrics: accuracy (overall correctness), precision (proportion of correct positive identifications), recall/sensitivity (ability to identify all relevant instances), specificity (correct identification of negatives), and F1 score (harmonic mean of precision and recall). However, these frame-based metrics inadequately capture the segmented nature of social interactions, where an algorithm producing rapid prediction alternations could achieve high F1 scores while generating ethologically meaningless results.

To address these limitations, we developed \textit{BANOS (Behavior Annotation Score)}, a set of metrics included in a package in \textit{Python}$^{\text{TM}}$ and \textit{MathWorks MATLAB}$^{\text{\textregistered}}$ that focuses on segment-level validation. \textit{BANOS} evaluates four key aspects: detection accuracy of behavioral segments, temporal overlap between predicted and ground truth segments, precision of start/end times, and intra-bout continuity \hyperref[box:banos]{(see Focus Box 2)}. This approach better respects the inherent nature of social interactions as continuous behavioral episodes rather than discrete frame classifications.

Beyond validating individual datasets, systematic tool evaluation requires benchmarking across standardized datasets. Few datasets for social interaction are publicly available with video or tracking data annotated by humans \cite{ref38,ref43,ref46,ref74,ref85,ref86,ref87,ref88}. Tools like \textit{DeepEthogram} \cite{ref38}, \textit{Keypoint-MoSeq} \cite{ref54}, \textit{SimBA} \cite{ref56}, and \textit{MARS} \cite{ref46} have undergone such benchmarking, though benchmark results are context-specific and may not generalize fully to new experimental conditions or behaviors not included in the benchmark datasets.

Two concepts distinguish effective tool evaluation: generalization measures how well tools maintain performance across different experimental conditions (arenas, lighting, mouse strains), while reproducibility determines whether behavioral patterns discovered in one dataset can be reliably detected in another. Both aspects are crucial for confirming that identified behaviors represent genuine biological phenomena rather than dataset-specific artifacts.

Traditionally, validation relies heavily on human interpretation, either through direct visualization or metrics that assess the replication of human annotations. Behaviors are deemed valuable based on human validation, which can be particularly challenging in unsupervised classification pipelines where outputs may not align with pre-existing human annotations or intuitive expectations. This could lead to the dismissal of behavior categories that might hold ethological or physiological significance for the studied animals, even if not immediately apparent to human observers. An alternative, less subjective approach validates behaviors by correlating them with physiological data, such as neuronal activity. High correlation between annotated behaviors and neuronal signals indicates biological relevance, independent of human judgment.

Several tools have been successfully applied to study neuronal correlates of social behavior. \textit{LiveMouseTracker} \cite{ref34}, \textit{3Dtracker} \cite{ref35}, \textit{SimBA} \cite{ref56}, and \textit{MARS} \cite{ref46} have revealed correlations with neuronal activity \cite{ref17,ref89,ref90,ref91,ref92,ref93,ref94,ref95}. \textit{CEBRA} \cite{ref72} and \textit{MARBLE} \cite{ref73} might also help uncover correlation patterns between neuronal activity and annotated behaviors, hinting at the neural encoding of these interactions. \textit{LISBET} \cite{ref63} further bridges computational behavior analysis and neurophysiology by identifying behavioral motifs correlated with neural activity in dopaminergic neurons of the Ventral Tegmental Area (VTA). This demonstrates the potential of computational annotations to reveal neurobiological significance beyond traditional human observations.

Establishing causality is paramount to definitively ascribe specific neuronal populations' roles in social behaviors, beyond mere correlations. While this is an expectation in neuroscientific studies, currently no tool provides the possibility to directly observe causality between neuronal populations and social interactions automatically detected. This is particularly complex as it often involves manipulating neuronal activity in real-time during behaviors, typically using optogenetic techniques. Notably, experiments outside of the \textit{OpenBehavior} platform references have been conducted where dopaminergic neurons of the Ventral Tegmental Area were stimulated lively to affect behaviors like resilience in fighting or promoting social behavior in groups of mice, demonstrating the potential for such approaches \cite{ref96,ref97}. Efforts remain to allow extended usage of these approaches by the community. Tools addressing human limitations and used to study neuronal activity, in the context of social behavior, are listed in \hyperref[tab:human_annotation_limitations]{Table 1}.

\section{Social interaction interpretation}

The final phase of the analysis interprets the validated data within the broader ethological and physiological contexts to draw meaningful conclusions about animal social interactions. This involves comparing the findings with existing research and integrating new observations with established scientific frameworks. Traditionally, the interpretation of outputs from statistical or machine learning analysis has been seen as a task only manageable by humans, due to the complex nature of translating analytical data into meaningful insights about social behavior while accounting for potential biases. This process demands a high level of intellectual and contextual engagement to bridge the gap between raw data and practical implications. However, the advent of large language models (LLMs) marks a significant shift in this perspective. LLMs, particularly those capable of multimodal interpretations such as \textit{GPT-4o} \cite{ref75}, are beginning to play a pivotal role in behavioral studies. These models are not only applying heuristic rules for classifying social interactions as illustrated before but are also advancing towards fully automating the analysis of social interactions. 

\onecolumn

% Table 3 - BANOS comparison
\begin{table*}[ht]
\vspace{0.5cm}
\centering
% \scriptsize
\setlength{\tabcolsep}{4pt}
\renewcommand{\arraystretch}{1.12}
\label{tab:banos_comparison}
\begin{tabular}{lcccc}
\toprule
\textbf{F1 Score} & \multicolumn{4}{c}{\textbf{BANOS}} \\
\textbf{(frame-based)} & \textbf{Detection Accuracy} & \textbf{Segment Overlap} & \textbf{Temporal Precision} & \textbf{Intra-Bout Continuity} \\
\midrule
0.79 & 0.27 & 0.15 & 0.01 & 0.76 \\
\bottomrule
\end{tabular}
\caption{\textbf{Comparison of agreement between human annotators on the \textit{CalMS21} dataset \cite{ref74} using the frame-based F1 score and the novel \textit{BANOS} metrics.} The low scores for Segment Overlap and Temporal Precision suggest that while humans largely agree on the presence of behaviors, they often disagree on the precise timing of these behaviors. All metrics range between 0 (false) and 1 (true).}
\end{table*}

\begin{focusbox}[breakable, enhanced, width=\textwidth]{Focus Box 3: Design considerations and trade-offs}
\label{box:design}

Choosing the right tools at the beginning of a research project is crucial, as these decisions profoundly influence the depth and scope of analysis and ultimately, the outcomes of the research. Often, researchers become attached to specific tools due to substantial initial investments. However, they may later encounter inherent limitations within these tools that restrict further analysis. It is essential to recognize these limitations early and understand the potential trade-offs to avoid the sunk cost fallacy and ensure that the selected tools are suitable for the intended purposes.

\vspace{0.5cm}

\textbf{Temporal Horizon and Latency}
\begin{itemize}[leftmargin=*,
                nosep,
                itemsep=4pt,
                parsep=0pt,
                label=\textbullet]
\item Considerations: Size of the time window for data analysis; using past/future frames to analyze the current one; input-output latency.
\item Trade-offs: Long windows provide more information and could improve classification performance, but they are more computationally expensive. Real-time analysis or neuronal computation can rely on past information only.
\item Implications: The latency is crucial for live monitoring scenarios, such as closed-loop experiments. The computational delay and the ability to deliver outputs solely depending on past data are critical factors in real-time experiments, while offline analysis allows for the use of future data frames and longer analyses to enhance predictions.
\end{itemize}

\vspace{0.5cm}

\textbf{Data Modality and Dimensionality}
\begin{itemize}[leftmargin=*,
                nosep,
                itemsep=4pt,
                parsep=0pt,
                label=\textbullet]
\item Considerations: Nature, dimensionality, and integration of various data types (e.g., video, depth sensing, RFID, physiological activity); experimental environment (e.g., home cage, arena).
\item Trade-offs: Multimodal and 3D video data enhance performance but increase costs and complexity. 2D video data is more accessible and easier to implement but less informative.
\item Implications: The data modality and dimensionality influence the depth of behavioral analysis, data acquisition complexity, and tool accessibility. 3D data captures subtle behavioral nuances but brings higher costs and computational demands, while 2D data is more accessible and sufficient for most studies but might lack details in specific analyses.
\end{itemize}

\vspace{0.5cm}

\textbf{Data Consistency and Diversity}
\begin{itemize}[leftmargin=*,
                nosep,
                itemsep=4pt,
                parsep=0pt,
                label=\textbullet]
\item Considerations: Quality of data used for training and testing.
\item Trade-offs: Refined, homogeneous data enhances performance but may misrepresent heterogeneous real-world conditions (e.g., varying data quality, cage dimensions, bedding).
\item Implications: Data consistency and diversity affect the robustness, generalization, and real-world applicability of the tool. Models developed in controlled environments may struggle when adapted to variable conditions and settings, such as fluctuations in lighting, cage sizes, or the physical conditions of the animals.
\end{itemize}

\vspace{0.5cm}

\textbf{Network Architecture and Hyperparameters}
\begin{itemize}[leftmargin=*,
                nosep,
                itemsep=4pt,
                parsep=0pt,
                label=\textbullet]
\item Considerations: Specific model architectures (e.g., transformer, LSTM, CNN) and their hyperparameters (e.g., number of layers, embedding dimensions, attention mechanisms).
\item Trade-offs: Different architectures have inherent strengths for certain problems—transformers excel at modeling long-range dependencies, CNNs at spatial patterns, and RNNs at sequential data. Hyperparameters impact model capacity, training stability, and computational requirements.
\item Implications: Architecture choices directly impact what patterns a model can detect and at what temporal resolutions. For example, transformers with attention mechanisms may better capture behaviors spanning multiple timescales, while temporal convolutional networks with dilated convolutions can efficiently model specific temporal receptive fields. These seemingly technical details fundamentally determine what types of social behaviors can be captured, how they're represented, and the model's ability to generalize across experimental conditions.
\end{itemize}

\vspace{1.2cm}

\textbf{Feature Engineering and Model Interpretability}
\begin{itemize}[leftmargin=*,
                nosep,
                itemsep=4pt,
                parsep=0pt,
                label=\textbullet]
\item Considerations: Selection of relevant features for behavioral analysis; explainability of the model's decisions.
\item Trade-offs: Human-engineered features are simple to interpret but may miss important information. Automated feature extraction drives discovery but can be difficult to interpret. The most performant model is not necessarily the easiest to explain.
\item Implications: Feature engineering and model interpretability impact performance, interpretability, discovery of novel insights, and user's trust. Researchers must balance innovation with clarity and reliability when deciding between manually selecting features or letting algorithms uncover new ones.
\end{itemize}

\vspace{0.5cm}

\textbf{Classification Paradigm}
\begin{itemize}[leftmargin=*,
                nosep,
                itemsep=4pt,
                parsep=0pt,
                label=\textbullet]
\item Considerations: Supervised learning and fine-tuning; unsupervised clustering.
\item Trade-offs: Supervised learning requires labeled data but aligns with human annotations. Unsupervised clustering allows for discovery but may not align with human understanding.
\item Implications: The classification paradigm affects the ease of model training, human intervention requirements, and the potential for discovering unforeseen behavioral clusters. Supervised learning might miss subtle yet significant behaviors, while unsupervised clustering could reveal undocumented behaviors at a higher computational cost and reduced interpretability.
\end{itemize}

\vspace{0.5cm}

\textbf{Output Granularity}
\begin{itemize}[leftmargin=*,
                nosep,
                itemsep=4pt,
                parsep=0pt,
                label=\textbullet]
\item Considerations: Level of output details; frame-by-frame vs. behavioral segment classification.
\item Trade-offs: Frame-by-frame classification may have higher performance but could lead to frequent label switching. Segment classification aligns better with human intuition but could lead to lower frame-level accuracy.
\item Implications: Output granularity dictates the balance between micro-level accuracy and macro-level behavioral insight, impacting the tool's practical relevance and usability.
\end{itemize}

\vspace{0.5cm}

\textbf{System Development, Deployment, and Maintenance}
\begin{itemize}[leftmargin=*,
                nosep,
                itemsep=4pt,
                parsep=0pt,
                label=\textbullet]
\item Considerations: Interplay between software and hardware prerequisites; time and resources needed for tool development.
\item Trade-offs: Customization boosts performance but increases development costs and setup complexity, potentially limiting accessibility and integration ease.
\item Implications: System development, deployment, and maintenance impact user accessibility, tool adoption, and adaptability to various practical scenarios, affecting overall deployment and performance optimization. Advanced tools may offer greater performance and deeper insights but require significant investments in technology and training, which may impact their accessibility and integration into broader research practices.
\end{itemize}

\vspace{0.5cm}

In summary, the process from data collection through the final application of analytical tools involves a complex sequence of decisions, each with profound implications. By integrating advanced data analysis techniques thoughtfully with practical deployment strategies, researchers can deepen their understanding of behavior and enhance the relevance and impact of their findings.
\end{focusbox}
\twocolumn

For instance, \textit{GPT-4o} \cite{ref75} demonstrated its capability by accurately interpreting the posture of mice and their relative positions from a single image prompted with a simple question. 
In another example, by analyzing three sequential images from a video, \textit{GPT-4o} \cite{ref75} could correctly classify the scene as a social agonistic event \hyperref[fig:llm_annotation]{(see Supplementary Figure 1)}. However, we should keep in mind that these new publicly available tools are not constrained to generate reproducible results for now, and their output should be carefully evaluated.

The potential of LLMs extends beyond automation of classification tasks; it enhances scientific understanding and aids in formulating new hypotheses. At the heart of this capability is the advanced representation learning of LLMs (i.e., extracting meaningful information from raw data) which is set to revolutionize research methodologies. These models promise to streamline labor-intensive tasks and substantially contribute to research by linking behavior with neuronal activity, potentially uncovering novel interpretations of social interactions, beyond human analytical capabilities. The integration of automated annotation tools with precise neurobiological techniques can offer deeper insights into the biological bases and causal relationships of behaviors. Furthermore, the coupling of unsupervised learning models with neurobiological methods may reveal new aspects of how social behavior is encoded in the brain, opening a new era for the understanding of brain functions and social behavior dynamics.

\section{Discussion}

In this work, we presented how the analysis of rodents' social behavior has started to change with the rise of machine learning (ML) and artificial intelligence (AI). We decomposed the typical analysis pipeline into a series of steps, providing references to the most established tools supporting each step. In doing so, we highlighted strengths and weaknesses of these tools or the underlying methods. Our hope is to help young researchers navigate the intricacies of social behavior analysis in the era of ML/AI, and experienced ones to engage in a constructive discussion on the development of the tools supporting the research community.

It is important to acknowledge that the field of social behavior analysis is rapidly evolving and what we recommend today may be superseded by more advanced tools in the near future. As new and powerful tools with diverse design aspects are continually emerging, researchers should stay informed about the latest developments. When incorporating tools within a research project, it is important to note that the functionalities and requirements of these tools are determined by the choices of their developers. Algorithms should be evaluated not only for their computational efficacy in synthetic benchmarks but also for how well they align with the specific research objectives. Crucial aspects beyond mere classification performance include temporal analysis and real-time performance, the modality and diversity of supported input data, feature extraction, model interpretability, and the granularity of outputs. These factors critically influence an algorithm's value in research studies, as detailed in \hyperref[box:design]{Focus Box 3}.

While many tools featured on community platforms like \textit{OpenBehavior} have facilitated access to automated behavior analysis, a number of innovative approaches not yet widely disseminated or easily usable offer conceptually powerful frameworks that could shape the future of social behavior research. Tools like \textit{LISBET} \cite{ref63}, \textit{SBeA} \cite{ref62}, \textit{DeepOF} \cite{ref61}, \textit{TREBA} \cite{ref79}, \textit{A-SOiD} \cite{ref77}, \textit{JABS} \cite{ref78}, and \textit{BAMS} \cite{ref64} each introduce distinctive concepts addressing key challenges in annotation scalability, temporal resolution, and behavioral discovery. Though these tools may currently lack user-friendly interfaces or broad accessibility, they exemplify important architectural and methodological advances (such as transformer-based self-supervision, programmatic expert input, and unsupervised motif discovery) that go beyond standard keypoint-to-class pipelines.

The development and adoption of tools in neuroscience are directly linked to the complexity of these tools and the accessibility to non-specialists, which can slow down their adoption and limit their use to a narrow group of experts. Furthermore, the diverse skills, methods, and terminologies used across disciplines can create barriers between neuroscientists, computer scientists, and behaviorists. The sustainability of these tools is further threatened by the temporary nature of academic positions. \hyperref[box:development]{Focus Box 4} explores these themes in greater detail, emphasizing the need for stable code, user-friendly interfaces, and robust documentation to ensure tools are accessible and maintainable long-term.

The transition from basic observational techniques to advanced, machine-assisted analyses show a significant evolution. Modern machine learning is particularly suited for handling large datasets and opens new avenues for real-time environmental adjustment in experiments. While the focus has traditionally been on rodents, the methodologies and technologies developed can be applied to more complex mammalian and even human social behaviors.

Despite the advancements, current techniques face several limitations. Most tools are designed for dyadic interactions rather than true group behavior analysis, which requires understanding complex social dynamics beyond multiple pairwise interactions. Additionally, tools often oversimplify complex social behaviors, categorizing them into distinct segments and focusing solely on a few specific behaviors. The reliability of these advanced tools depends critically on the quality and diversity of the datasets they are trained on, and on their ability to handle common tracking issues like jittering, identity swaps, and missing keypoints. Understanding the specific contexts and applications of diverse datasets is crucial to prevent misuse. Comprehensive and diverse datasets are necessary for objective analysis, capturing the full spectrum of rodent behaviors.

In recent years, the open release of social behavior dataset used for benchmarking \cite{ref38,ref43,ref46,ref64,ref74,ref85,ref86,ref87,ref88} has greatly contributed to the development of better analysis tools and more fair evaluation of the existing ones \cite{ref38,ref46,ref54,ref63,ref98}. However, over-reliance on measures of classification performance can introduce undesirable biases. Notably, algorithms that surpass the typical level of agreement between human annotators might reflect overfitting to a specific annotator's style, rather than capturing a broader consensus or accurately representing the behaviors under investigation. 

\onecolumn
\begin{focusbox}[breakable, enhanced]{Focus Box 4: Tool development, maintenance and adoption}
\label{box:development}
In the last decade we experienced the rapid growth of specialized tools for behavioral analysis, and more recently the trend has emerged specifically for social behavior \hyperref[fig:development]{(see Figure 5a)}. However, these tools face slow adoption due to their complexity, limiting use to specialists and posing challenges for biologists. We quantified this trend by categorizing their citations as “technical”, “research”, or “review”, and observed a marked relative decrease in research and biological applications compared to technical articles over the past decade \hyperref[fig:development]{(see Figure 5b)}. This finding highlights how these tools are still not systematically used to answer biological questions. This observation should not be interpreted as evidence that such tools are limited in value, or destined to remain unused. Rather, it highlights the inherent challenges posed by an academic system that does not always provide clear incentives or frameworks for transitioning novel techniques into mainstream biological research. The slower uptake is therefore less a consequence of the tools’ shortcomings or their developers’ efforts, and more a reflection of the structural hurdles leading to lack of support or resources to integrate these methods into research.

\vspace{0.5cm}

Furthermore, conceptual breakthroughs often begin as proofs of concept: generating innovative ideas and methods (even if they are not immediately adopted) is a necessary and pivotal part of scientific progress. While some projects excel at translating tools across the entire pipeline from design to large-scale dissemination, others focus primarily on introducing novel concepts. Both approaches are essential. By producing a diverse “pool” of ideas, we foster opportunities for future exploration and eventual application. In time, these currently specialized or underutilized tools can evolve (or be integrated by others) to meet broader needs, thereby driving the field forward in new and unexpected ways.

\vspace{0.5cm}

Collaborations between neuroscientists, computer scientists, and behaviorists have been fruitful historically. Yet, the machine learning and neuroscientific communities often find themselves at a linguistic and conceptual impasse with different skills, methods and terminology. There is a shortage of professionals skilled in observational and computational neuroscience, highlighting the need for interdisciplinary education to cultivate researcher’s adept in both fields. 

\vspace{0.5cm}

% Figure 5 - Full width
\begin{figure}[H]
\centering
\includegraphics[width=0.9\textwidth]{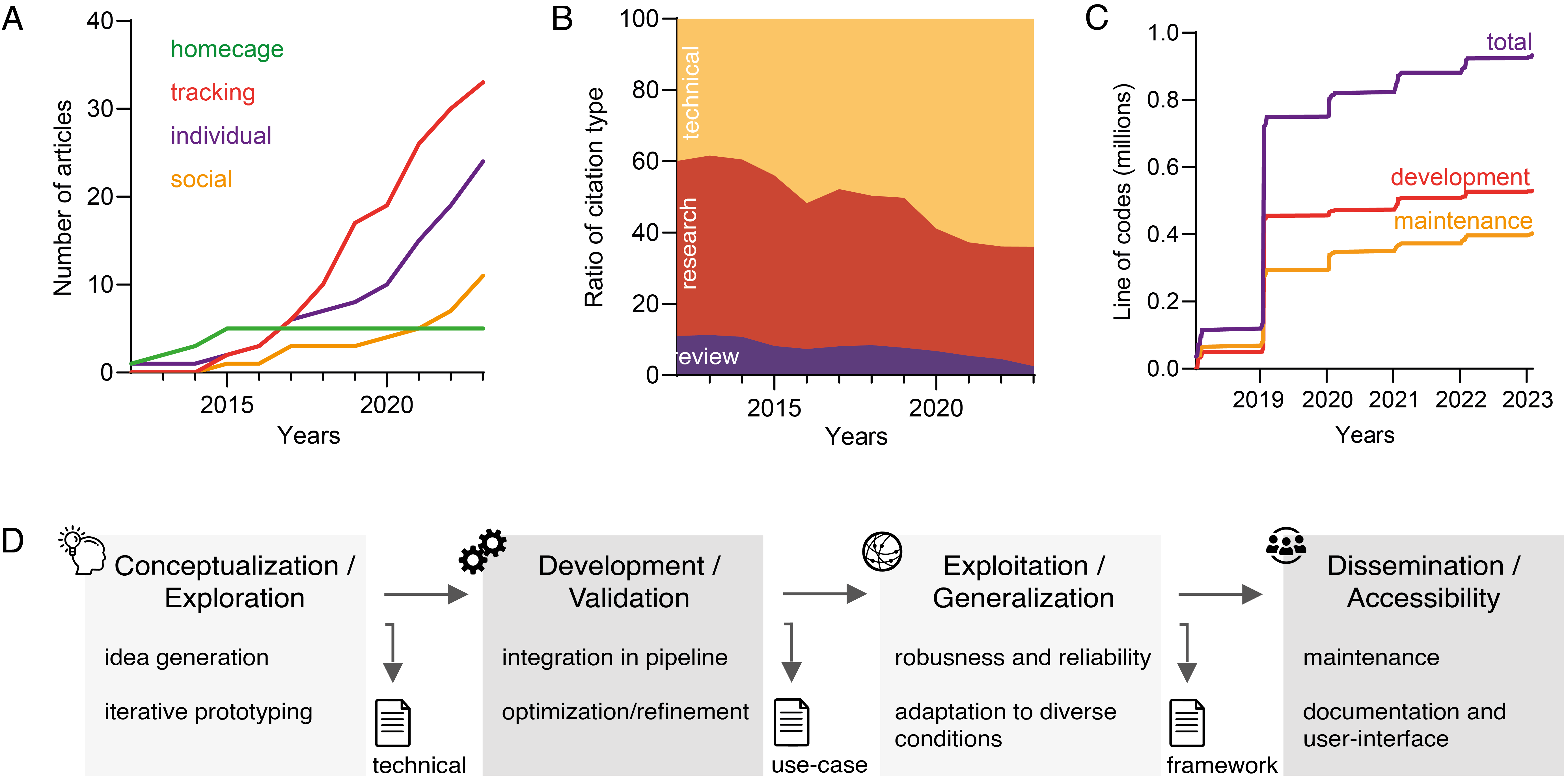}
\caption{\textbf{Tools development, maintenance and adoption.} \textbf{a}. Cumulative number of articles per year by category. The development of tools follows a rapid increase but categories such as social behavior analysis remain underrepresented. \textbf{b}. Relative number of citations per year across three groups (technical, research, or review). The trend shows a decrease in research and biological applications compared to technical articles over the past decade, possibly indicating the community's challenges in applying these tools in contrast with the rapid development of new ones. This classification was based on titles and abstracts from PubMed citing the tools referenced, analyzed using OpenAI GPT-4 API \cite{ref75}. Out of 2,943 citations, we manually checked a sample of 50 citations to validate the performance of this automatic classification. \textbf{c}. Cumulative lines of code committed per year in DeepLabCut's GitHub repository (https://github.com/DeepLabCut/DeepLabCut). The replaced lines of code served as a proxy to assess maintenance efforts. \textbf{d}. Developmental stages of analytical tools, each with distinct goals, outputs, and community impacts. Stages include conceptualization and technical proof (valued through technical papers), integration into applied research pipelines (use-case papers), and generalization of exploitation across conditions (framework papers). However, contributions in dissemination, accessibility, and usability are not typically valued through traditional academic channels or publications. Data used in \textbf{a}, \textbf{b} and \textbf{c} is relative to tools referenced in \hyperref[tab:tool_list]{Supplementary Table 1}. We selected tools referenced in \textit{OpenBehavior} as a strategy to focus on those likely intended for community use by their authors.}
\label{fig:development}
\end{figure}

\vspace{0.5cm}

Effective tool development requires bridging the gap between complex biological research and practical tool application, necessitating a blend of diverse expertise. However, ensuring that tools are independently usable across various labs, and providing necessary support, presents significant challenges. The temporary nature of postdoctoral positions, typically held by tool authors, risks knowledge loss, threatening the long-term sustainability of these tools within a single lab. Furthermore, the importance of stable Application Programming Interfaces (APIs), user-friendly interfaces and comprehensive documentation cannot be overstated, alongside policies promoting not only widespread release but also usability of these tools. DeepLabCut \cite{ref40}, SLEAP \cite{ref42}, and SIMBA \cite{ref56} are rare examples of successful tools overcoming these barriers. \hyperref[fig:development]{(Figure 5c)} illustrates the cumulative lines of code committed per year in DeepLabCut's GitHub repository, reflecting the workload of development and maintenance \hyperref[fig:development]{(see Figure 5c)}. Here we consider the replacement of lines of code as a proxy to assess maintenance efforts and the sustained demand of tool upkeep.

\vspace{0.5cm}

Tool development can be conceptualized as a multi-stage effort. These stages are critical for transforming initial technological concepts into practical and widely applicable research tools.

\begin{itemize}[leftmargin=*,
                nosep,
                itemsep=2pt,
                parsep=0pt,
                label=\textbullet]
\item Conceptualization and Technical Proof: This initial stage involves the theoretical development of tools and their first practical implementations. Innovations are often tested in controlled environments to prove their feasibility. Technical papers primarily document these early proofs, focusing on the technical capabilities and potential applications of the tool.
\item Integration into Applied Research Pipelines: Once tools have demonstrated their technical viability, the next phase involves integrating them into actual research settings. Use-case papers emerge during this stage, detailing the application of these tools in specific studies and illustrating their effectiveness in capturing and analyzing data under varied real research conditions. 
\item Generalization of Tool Exploitation Across Conditions: The final stage of development seeks to generalize the use of the tool across various research conditions and settings, demonstrating its robustness and flexibility. Framework papers typically emerge here, outlining standardized methodologies for employing these tools in a broad range of scientific inquiries. This stage is crucial for establishing the tool as a reliable and essential instrument in the scientific community.
\end{itemize}

\vspace{0.5cm}

\hyperref[fig:development]{(Figure 5d)} illustrates these stages in the life cycle of tool development, emphasizing how tools mature from conceptual models into integral components of the research infrastructure \hyperref[fig:development]{(see Figure 5d)}. Despite its importance, contributions in dissemination, accessibility, and usability are often underrepresented in traditional academic outputs. However, these aspects are vital for ensuring that tools not only meet the technical specifications but also are accessible and usable by a broader segment of the research community, including those who may not have specialized training in computational methods. This requires dedication to maintenance as well as proper documentation and creation of user-interfaces, which is not yet sufficiently recognized in academia as it does not translate to publications. Furthermore, the research community frequently overlooks the sustained costs associated with tool maintenance, and there is an ongoing shortage of funding mechanisms dedicated to supporting these expenses over the long term. While open-source projects demonstrate the effectiveness of community-driven development, their long-term sustainability often depends on entities with sufficient resources. Commercial entities have established a sustainable model by offering specialized functions, updates, and support, but maintaining a balance between commercial interests and academic integrity is crucial \cite{ref102}. Combining open-source and commercial approaches can foster collaboration and ensure ongoing maintenance. Among other possibilities, this can be achieved through models like dual licensing (where software can be released both under an open-source license and a commercial license), open-core (where the core software is open-source, but additional premium features and extended support are offered commercially), or partnerships where commercial entities sponsor open-source projects without demanding changes that compromise the project’s goals.

\vspace{0.5cm}

For researchers keen on adopting these tools, selecting algorithms that are well-maintained or based on stable functions is advisable. This ensures longevity and alignment with future data collection needs. Such an approach not only leverages the strengths of existing algorithms but also remains open to emerging technological developments, fostering a collaborative and interdisciplinary research environment.

\end{focusbox}
\twocolumn

We tested this hypothesis on the \textit{CalMS21} dataset, instrumental and widely used to benchmark algorithms \cite{ref74}. By comparing video samples annotated by multiple annotators using the F1 score, a typical measure of algorithmic performance, we quantified the level of human agreement on the dataset (F1 = 0.79). This value can be considered as the limit of the relevance of an algorithm on the \textit{CalMS21} dataset, as it corresponds to the average consensus between different human annotators. Therefore, we argue that a balanced and critical approach to dataset evaluation and algorithm development is essential in rodent behavior research.

In an era dominated by automated data processing, human oversight remains essential. The tendency of deep learning methods to act as "black boxes" underscores the importance of maintaining a critical perspective on how computational tools are developed and employed. By integrating human insights with machine learning and fostering interdisciplinary collaborations, we can enhance the validity and applicability of our research findings. The combined use of advanced machine learning tools and detailed human analysis promises to significantly improve our understanding of the neurobiological correlates of social behavior. This collaborative impact could revolutionize approaches to mental health treatment and inform social policies, marking a new frontier in behavioral science.

% Acknowledgments
\section*{Acknowledgments}
We would like to thank C. Lüscher, A. Benjamin and S. El Boustani for comments on the manuscript.

This work was supported by the European Research Council (ERC SocialNAc 864552), the Fondation HUG and the Swiss National Science Foundation (SNSF 310030-212219).

% Code availability
\section*{Code availability}
The \textit{Behavior Annotation Scores (BANOS)}, featuring \textit{Python}$^{\text{TM}}$ and \textit{MathWorks MATLAB}$^{\text{\textregistered}}$ implementations, is available in the following repository: https://github.com/BelloneLab/BANOS.

% This would typically be generated from .bib file
\bibliographystyle{unsrtnat}
% \bibliography{references}

\clearpage
\onecolumn
\appendix
\section*{Supplementary Material}
\renewcommand{\thefigure}{S\arabic{figure}}  % Use S1, S2, ... for Supplementary Figures
\setcounter{figure}{0}  % Reset figure counter
\renewcommand{\thetable}{S\arabic{table}}  % Use S1, S2, ... for Supplementary Figures
\setcounter{table}{0}  % Reset figure counter

\begin{figure}[H]
\centering
\includegraphics[width=0.9\textwidth]{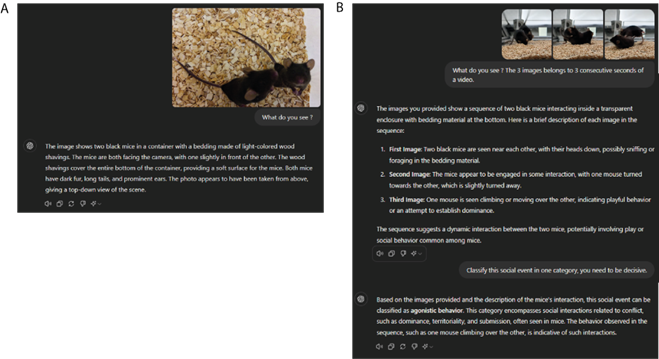}
\caption{
        \textbf{Demonstration of the ability of ChatGPT-4o \cite{ref75} to interpret mouse social interactions from (a) a single image and (b) a sequence of three images, based on user prompts.} \textbf{a.} Single image with prompt and output of ChatGPT-4o \cite{ref75} to assess capacity to interpret social posture between mice. \textbf{b.} Three images with prompts and outputs of ChatGPT-4o \cite{ref75} to assess capacity to interpret social interaction between mice.
    }
\label{fig:llm_annotation}
\end{figure}

\clearpage
\begin{figure}[H]
\centering
\includegraphics[width=0.6\textwidth]{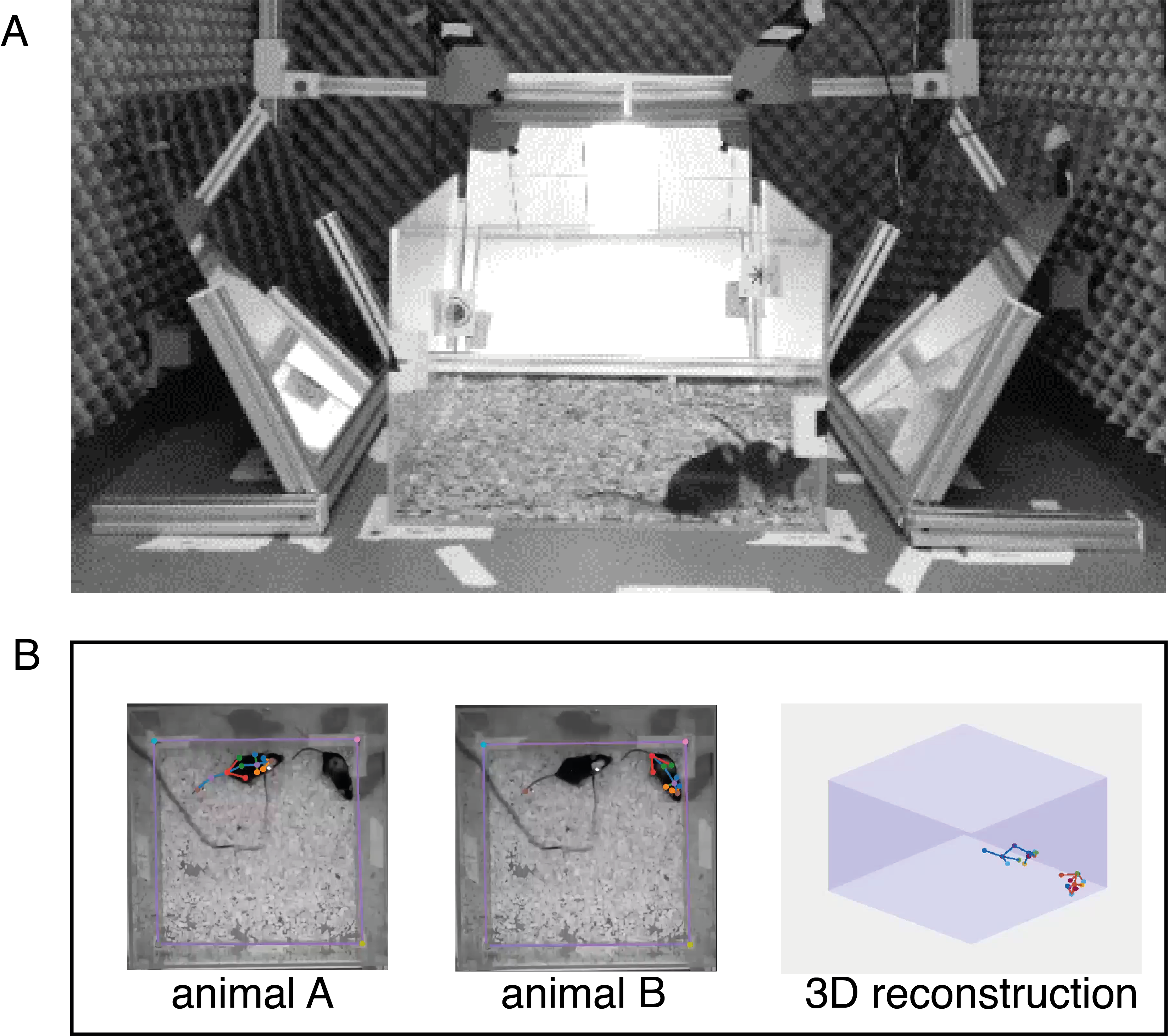}
\caption{
        \textbf{Illustration of a data acquisition setup using a single camera and mirrors for 3D pose reconstruction.} \textbf{a.} Example of recording setup using camera and mirrors. \textbf{b.} Left and middle, tracking of animal A and B using \textit{DeepLabCut} \cite{ref40}. Right, 3D reconstruction using \textit{Anipose} \cite{ref52}.
    }
\label{fig:3d_acquisition}
\end{figure}

\clearpage
\begin{table}[ht]
    \centering
    \scriptsize
    \renewcommand{\arraystretch}{1.15}
    \setlength{\tabcolsep}{4pt}
    \label{tab:tool_list}
    \begin{tabular*}{0.7\textwidth}{@{\extracolsep{\fill}} L{3.3cm} C{0.9cm} C{0.9cm} C{0.9cm} C{0.9cm} L{3.5cm} }
        \toprule
        \textbf{Tool} & \textbf{Homecage} & \textbf{Tracking} & \textbf{Individual} & \textbf{Social} & \textbf{References} \\
        \midrule
        LocoMouse               & \cmark & \dash  & \cmark & \dash  & \cite{ref103} \\
        JABS                   & \cmark & \dash  & \cmark & \dash  & \cite{ref78} \\
        Eco-HAB               & \cmark & \dash  & \dash  & \dash  & \cite{ref33} \\
        Phenopy               & \cmark & \dash  & \dash  & \dash  & \cite{ref104} \\
        LiveMouseTracker      & \cmark & \dash  & \dash  & \dash  & \cite{ref34} \\
        AutonoMouse           & \cmark & \dash  & \dash  & \dash  & \cite{ref105} \\
        3DTracker             & \dash  & \cmark & \dash  & \cmark & \cite{ref35} \\
        MouseMove             & \dash  & \cmark & \dash  & \dash  & \cite{ref106} \\
        ToxTrac               & \dash  & \cmark & \dash  & \dash  & \cite{ref47} \\
        OptiMouse             & \dash  & \cmark & \dash  & \dash  & \cite{ref107} \\
        DeepLabCut            & \dash  & \cmark & \dash  & \dash  & \cite{ref40,ref108,ref109} \\
        ART                   & \dash  & \cmark & \dash  & \dash  & \cite{ref110} \\
        MouBeAT               & \dash  & \cmark & \dash  & \dash  & \cite{ref111} \\
        Picamera              & \dash  & \cmark & \dash  & \dash  & \cite{ref112} \\
        idtracker.ai          & \dash  & \cmark & \dash  & \dash  & \cite{ref113} \\
        ezTrack               & \dash  & \cmark & \dash  & \dash  & \cite{ref114} \\
        DeepBehavior          & \dash  & \cmark & \dash  & \dash  & \cite{ref41} \\
        DeepPoseKit           & \dash  & \cmark & \dash  & \dash  & \cite{ref115} \\
        LEAP                  & \dash  & \cmark & \dash  & \dash  & \cite{ref116} \\
        RAT                   & \dash  & \cmark & \dash  & \dash  & \cite{ref117} \\
        TRex                  & \dash  & \cmark & \dash  & \dash  & \cite{ref49} \\
        DeepLabStream         & \dash  & \cmark & \dash  & \dash  & \cite{ref118} \\
        Anipose               & \dash  & \cmark & \dash  & \dash  & \cite{ref52} \\
        LiftPose3D            & \dash  & \cmark & \dash  & \dash  & \cite{ref53} \\
        SLEAP                 & \dash  & \cmark & \dash  & \dash  & \cite{ref42} \\
        FreiPose              & \dash  & \cmark & \dash  & \dash  & \cite{ref119} \\
        PyMouseTracks         & \dash  & \cmark & \dash  & \dash  & \cite{ref120} \\
        Lightning Pose        & \dash  & \cmark & \dash  & \dash  & \cite{ref121} \\
        ETHOWATCHER           & \dash  & \cmark & \cmark & \dash  & \cite{ref122} \\
        M-Track               & \dash  & \cmark & \cmark & \dash  & \cite{ref123} \\
        CaT-z                 & \dash  & \cmark & \cmark & \dash  & \cite{ref124} \\
        EthoLoop              & \dash  & \cmark & \cmark & \dash  & \cite{ref125} \\
        DANNCE                & \dash  & \cmark & \cmark & \dash  & \cite{ref43} \\
        CAPTURE               & \dash  & \cmark & \cmark & \dash  & \cite{ref126} \\
        AnimalTA              & \dash  & \cmark & \cmark & \dash  & \cite{ref50} \\
        Hong Workflow         & \dash  & \cmark & \cmark & \cmark & \cite{ref36} \\
        MoSeq                 & \dash  & \cmark & \cmark & \cmark & \cite{ref57} \\
        SIPEC                 & \dash  & \cmark & \cmark & \cmark & \cite{ref51} \\
        3DDD Social Mouse Tracker & \dash & \cmark & \cmark & \cmark & \cite{ref37} \\
        AlphaTracker          & \dash  & \cmark & \cmark & \cmark & \cite{ref45} \\
        MotionMapper          & \dash  & \dash  & \cmark & \dash  & \cite{ref60} \\
        JAABA                 & \dash  & \dash  & \cmark & \dash  & \cite{ref127} \\
        Pathfinder            & \dash  & \dash  & \cmark & \dash  & \cite{ref128} \\
        PAWS                  & \dash  & \dash  & \cmark & \dash  & \cite{ref129} \\
        B-SOiD                & \dash  & \dash  & \cmark & \dash  & \cite{ref58} \\
        DeepEthogram          & \dash  & \dash  & \cmark & \dash  & \cite{ref38} \\
        PS-VAE                & \dash  & \dash  & \cmark & \dash  & \cite{ref130} \\
        VAME                  & \dash  & \dash  & \cmark & \dash  & \cite{ref59} \\
        BehaviorDEPOT         & \dash  & \dash  & \cmark & \dash  & \cite{ref131} \\
        SEB3R                 & \dash  & \dash  & \cmark & \dash  & \cite{ref132} \\
        Keypoint-MoSeq        & \dash  & \dash  & \cmark & \cmark & \cite{ref54} \\
        SimBA                 & \dash  & \dash  & \dash  & \cmark & \cite{ref56} \\
        MARS                  & \dash  & \dash  & \dash  & \cmark & \cite{ref46} \\
        \bottomrule
    \end{tabular*}
    \caption{
        \textbf{ List of rodent behavior analysis tools referenced on the \textit{OpenBehavior} platform (https://edspace.american.edu/openbehavior) as of May 2024.} Tools are categorized by their suitability for homecage monitoring, tracking, and analyzing individual or social behaviors.
    }
\end{table}

\end{document}